\documentclass[final]{ecai}

\usepackage{latexsym}
\usepackage{amssymb}
\usepackage{amsmath}
\usepackage{cleveref}
\usepackage{amsthm}
\usepackage{booktabs}
\usepackage{enumitem}
\usepackage{graphicx}
\usepackage{color}
\usepackage{acronym}
\acrodef{LLM}{Large Language Model}

\usepackage{subcaption}
\usepackage[flushleft]{threeparttable}
\usepackage{float}

\newcommand{\BibTeX}{B\kern-.05em{\sc i\kern-.025em b}\kern-.08em\TeX}

\begin{document}

\begin{frontmatter}

\paperid{123}

\title{Teuken-7B-Base \& Teuken-7B-Instruct: Towards European LLMs}

\author{
\normalfont 
Mehdi Ali\textsuperscript{1,2}$^\dagger$,
Michael Fromm\textsuperscript{1,2}$^\dagger$,
Klaudia Thellmann\textsuperscript{3,8}$^\dagger$, 
Jan Ebert\textsuperscript{4}$^\dagger$, 
Alexander Arno Weber\textsuperscript{1,2}$^\dagger$,\\
Richard Rutmann\textsuperscript{1,2},
Charvi Jain\textsuperscript{1,2},
Max Lübbering\textsuperscript{1,2},
Daniel Steinigen\textsuperscript{1},
Johannes Leveling\textsuperscript{1}, \\
Katrin Klug\textsuperscript{1},
Jasper Schulze Buschhoff\textsuperscript{1},
Lena Jurkschat\textsuperscript{3},
Hammam Abdelwahab\textsuperscript{1}, \\
Benny Jörg Stein\textsuperscript{1},
Karl-Heinz Sylla\textsuperscript{1},
Pavel Denisov\textsuperscript{1},
Nicolo' Brandizzi\textsuperscript{1},
Qasid Saleem\textsuperscript{1}, \\
Anirban Bhowmick\textsuperscript{1}, 
Lennard Helmer\textsuperscript{1}, 
Chelsea John\textsuperscript{4},
Pedro Ortiz Suarez\textsuperscript{5},
Malte Ostendorff\textsuperscript{5},
Alex Jude\textsuperscript{1}, \\
Lalith Manjunath\textsuperscript{3},
Samuel Weinbach\textsuperscript{7},
Carolin Penke\textsuperscript{4},
Oleg Filatov\textsuperscript{4},
Fabio Barth\textsuperscript{5}, \\
Paramita Mirza\textsuperscript{6},
Lucas Weber\textsuperscript{6},
Ines Wendler\textsuperscript{1},
Rafet Sifa\textsuperscript{1},  
Fabian Küch\textsuperscript{6},
Andreas Herten\textsuperscript{4},
René Jäkel\textsuperscript{3}, 
Georg Rehm\textsuperscript{5}, 
Stefan Kesselheim\textsuperscript{4},
Joachim Köhler\textsuperscript{1},
Nicolas Flores-Herr\textsuperscript{1} \\
\bigskip
\textsuperscript{1}Fraunhofer IAIS, 
\textsuperscript{2}Lamarr Institute,
\textsuperscript{3}TU Dresden,
\textsuperscript{4}FZ Jülich, 
\textsuperscript{5}DFKI, 
\textsuperscript{6}Fraunhofer IIS, 
\textsuperscript{7}Aleph Alpha,
\textsuperscript{8}ScaDS.AI Dresden/Leipzig
\thanks{\textdagger Equal contribution.}}

\begin{abstract}
We present two multilingual LLMs, \emph{Teuken 7B-base} and \emph{Teuken 7B-instruct}, designed to embrace Europe's linguistic diversity by supporting all 24 official languages of the European Union. 
Trained on a dataset comprising around 60\% non-English data and utilizing a custom multilingual tokenizer, our models address the limitations of existing \acp{LLM} that predominantly focus on English or a few high-resource languages. 
We detail the models' development principles, i.e., data composition, tokenizer optimization, and training methodologies. 
The models demonstrate strong performance across multilingual benchmarks, as evidenced by their performance on European versions of ARC, HellaSwag, and TruthfulQA.
\end{abstract}

\end{frontmatter}

\section{Introduction}
\acp{LLM} represents a disruptive technology that has the potential to be applied in numerous applications. 
To develop \acp{LLM}, expertise in various areas is required, starting from large-scale data pre-processing~\cite{DBLP:journals/corr/abs-2406-17557}, training efficient tokenizers~\cite{DBLP:conf/nips/PetrovMTB23, ali-etal-2024-tokenizer}, pre-training the models efficiently on a vast infrastructure spanning thousands of GPUs, instruction tuning the pre-trained models, and properly evaluating them on various downstream tasks~\cite{dubey2024llama3herdmodel}.
Multilingualism as an additional dimension introduces additional considerations to all phases of the model development, such as multilingual data composition and multilingual tokenizer.
Therefore, it is crucial that the technology and expertise to build these models is democratized to enable different communities and organizations to employ these models for their use cases.

Many efforts in developing open-source \acp{LLM} have been undertaken, such as BLOOM~\cite{DBLP:journals/corr/abs-2211-05100}, LLaMA-3~\cite{dubey2024llama3herdmodel}, OLMo~\cite{DBLP:conf/acl/GroeneveldBWBKT24}, Aya~\cite{DBLP:conf/acl/UstunAYKDOBSOKV24}, and Mistral~\cite{DBLP:journals/corr/abs-2310-06825}.

While these existing efforts represent significant contributions to the community and the advancement of artificial intelligence, there are still two major limitations.
First, the current open-source models are predominantly English-centric, limiting their use to a broad multilingual context covering high- and low-resource languages such as within the European Union.
Relying on English-centric models that employ an English-centric tokenizer in a multilingual context introduces severe disadvantages, such as lower downstream performance, additional inference costs, and increased training costs for language-specific continued pre-training and fine-tuning for languages besides English~\cite{DBLP:conf/nips/PetrovMTB23,ali-etal-2024-tokenizer}.
Second, open-source efforts disclose different levels of granularity in sharing details about the development of the models.
Although information on the model architecture is usually shared, the decisions/ablation experiments behind certain architectural choices are not always described, which can provide crucial insights relevant to developing custom \acp{LLM}.
This limitation is even more apparent in the description of the dataset composition, which often provides a coarse overview, hampering the reproduction of the work.

To address the aforementioned limitations, we created a multilingual base model that has been trained on top of all 24 European official languages and the corresponding instruction-tuned model.
In particular, we make the following contributions:

\begin{itemize}
    \item We present Teuken-7B-Base\footnote{\url{https://huggingface.co/openGPT-X/Teuken-7B-base-v0.6}}, a European pre-trained \ac{LLM} that has been trained from scratch for 6 trillion tokens based on all 24 official European languages.
    \item We present Teuken-7B-Instruct\footnote{\url{https://huggingface.co/openGPT-X/Teuken-7B-instruct-v0.6}}, the instruction-tuned model on top of the base model to enhance instruction-following capabilities.
\end{itemize}

In addition to these artefacts, we describe our design decisions in detail to facilitate the reproduction of our work and the development of novel models based on it.
Therefore, our contributions present an essential step towards the democratization of this technology across Europe.

\section{Related Work}

Since the introduction of GPT-3~\cite{DBLP:conf/nips/BrownMRSKDNSSAA20}, several open-source/open-weights efforts have been undertaken to train \acp{LLM}. 
While the large majority of the work focus on English-centric models~\cite{DBLP:journals/corr/abs-2205-01068,DBLP:journals/corr/abs-2302-13971,DBLP:journals/corr/abs-2307-09288,DBLP:conf/acl/GroeneveldBWBKT24,dubey2024llama3herdmodel}, there have been also efforts training multilingual models.

One of the most prominent examples is BLOOM~\cite{DBLP:journals/corr/abs-2211-05100}, a 176B \ac{LLM} trained on 46 natural languages. 
Further examples that specifically address multilingualism are the encoder-decoder models mT5~\cite{DBLP:conf/naacl/XueCRKASBR21}, XLM~\cite{lample2019crosslinguallanguagemodelpretraining}, XLM-R~\cite{conneau-etal-2020-unsupervised}, and the encoder model mBERT~\cite{devlin-etal-2019-bert}.

Unlike the previously mentioned efforts, we specifically address 24 official European languages and ensure that a significant fraction of the training data comprises non-English data, representing a major step towards European \acp{LLM}.
Concurrent to our work, EuroLLM~\cite{martins2024eurollm}, a 1.7B and 9B decoder-only \ac{LLM} that follows the same spirit as our undertaking by addressing all 24 European languages and Salamandra\footnote{\url{https://huggingface.co/BSC-LT/salamandra-7b}} covering 32 languages, have been presented.

\section{Pre-Training Data}\label{section:data}

We used the dataset presented in~\cite{br2024data} as a base since it i.) covers all 24 official EU languages, ii.) is large, iii.) contains a large amount of non-English data, iv.) comprises various domains, and v.) is filtered based on established practices.
In the following, we describe the composition of our final training comprised of web-crawled and curated data.

For the web crawled part, we sampled our dataset based on the 60 filtered Common Crawls WET dumps (cf. ~\Cref{sec:dumps}) provided by Brandizzi \textit{et al.}~\cite{br2024data}.
The raw WET dumps have been filtered based on established heuristics, ensuring that the pre-training corpus is of high quality~\cite{br2024data}. 
Additionally, we employed the recently released Fine-Web EDU~\cite{DBLP:journals/corr/abs-2406-17557} and DCLM~\cite{li2024datacomp} dataset since significant performance improvements for models trained on these datasets have been reported.
It should be noted that Fine-Web EDU is only available in English. 
We still integrated it and investigated the cross-lingual performance of the corresponding model checkpoint (see \Cref{section:model_training}).
To ensure a multilingual composition, we up-sampled all languages except English, which we down-sampled.

In addition to web crawled data, we included all curated datasets from~\cite{br2024data}.
Part of the curated datasets are domain-specific data such as academics, finance, and patents ensuring the diversity of our training dataset.

\begin{figure*}[h!]
    \centering
    \includegraphics[width=0.95\linewidth]{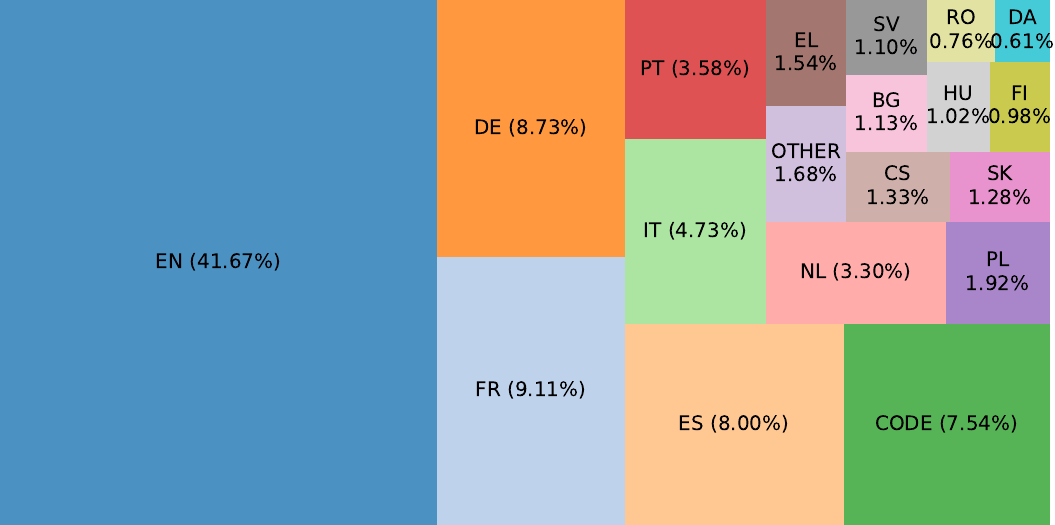}
    \caption{Language distribution of the tokenized dataset, comparing the presence of English and other European languages.}
    \label{fig:treemap_main}
\end{figure*}

\begin{figure*}[h!]
    \centering
    \includegraphics[clip, trim=0 0.7cm 0 0.6cm, width=0.95\linewidth]{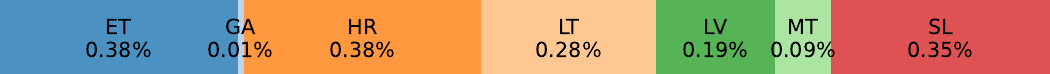}
    \caption{Breakdown of the ``OTHER'' category.}
    \label{fig:treemap_other}
\end{figure*}

Our composed training dataset contains 6 trillion tokens, of which 86.79\% (cf.~\Cref{tab:combined_words}) originates from web data, and the remaining 13.21\% represent is curated data. 
As illustrated in~\Cref{fig:treemap_main} and~\Cref{fig:treemap_other}, 41.70\% of the tokens stem from English content and due to the inclusion of German, French, and Spanish, we approach around two-thirds of the total tokens.

\section{Multilingual Tokenization and Fertility Impact on Model Efficiency}

In multilingual natural language processing (NLP), it is crucial to train balanced multilingual tokenizers~\cite{DBLP:conf/nips/PetrovMTB23,ali-etal-2024-tokenizer} to avoid 
increased training and inference costs and latency during inference for non-English queries.
Furthermore, it prevents the model from learning long-range dependencies in limited context windows~\cite{DBLP:conf/nips/VaswaniSPUJGKP17}.

Therefore, we developed a custom multilingual tokenizer, closely following \citet{ali-etal-2024-tokenizer}, that is optimized for all 24 official European languages. The tokenizer training dataset contains an equal number of documents for each of the 24 languages. The documents were sourced from the same data as used in pretraining.
It aims to reduce excessive text fragmentation, a phenomenon termed high ``fertility'', and refers to the average number of tokens generated per word. 

Fertility (\( F \)) is defined as the ratio of the total number of tokens (\( T \)) to the total number of words (\( W \)) in a text, i.e., $F = \frac{T}{W}$.

\begin{figure}[h!]
    \centering    \includegraphics[width=0.9\linewidth]{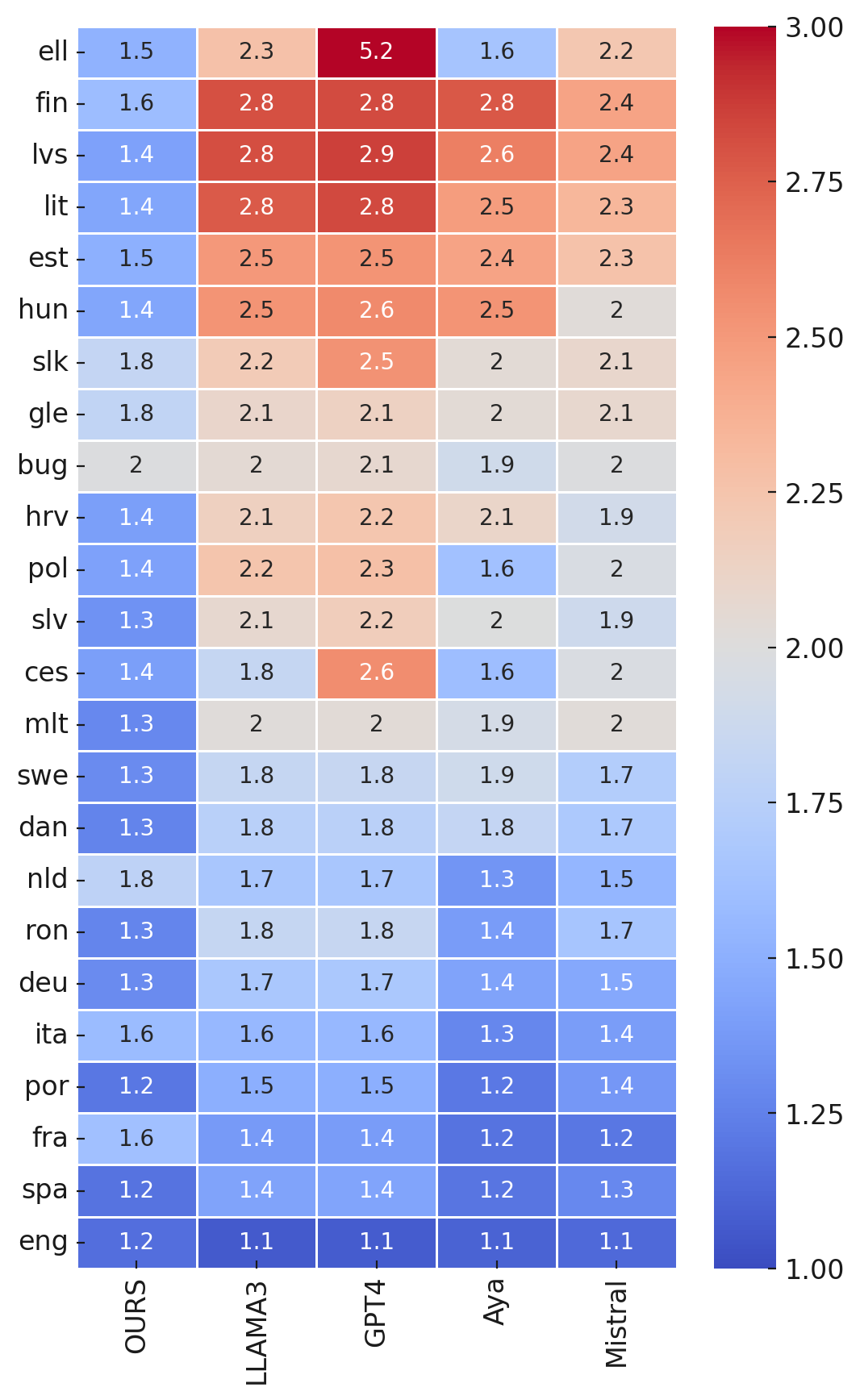}
    \caption{Fertility across the official 24 European languages.}
    \label{fig:fertility}
\end{figure}

We conducted a fertility analysis on 2,000 sentences from the FLORES-200~\cite{DBLP:journals/corr/abs-2207-04672} dataset to compare tokenizers. 
Because the dataset is translated across languages, i.e., the analysis is conducted on semantic equivalent content, it provides a reliable basis for evaluation. 
A comparison with other widely-used tokenizers is presented in~\Cref{fig:fertility}

Our custom tokenizer demonstrates that for 19 out of the 24 languages, fertility values are similar or lower than those of related tokenizers. 
This effect is especially pronounced in languages with complex morphology or long word structures, such as Finnish, German, and Hungarian. 

Lowering fertility enables longer queries and documents to be processed without exceeding the context window. 
This is particularly advantageous in tasks that require the processing of legal or medical documents, where maintaining the integrity of long documents is essential for accurate understanding.

\section{Base Model}\label{section:base_model}

In the following, we describe the model architecture (\Cref{section:model_architecture}) and training (\Cref{section:model_training}).
Additionally, we describe in the appendix the used training framework (\Cref{section:software}) and the training infrastructure  (\Cref{section:training_infrastructure}).

\subsection{Model Architecture}\label{section:model_architecture}

Our model is a 7B transformer-based decoder-only model.
\Cref{table:model_hyper_parameters} provides an overview of our model architecture.
We want to highlight that our architectural choices are derived from internal ablation studies and findings from related work. 
Our models have a sequence length of 4096 tokens and employ Rotary~\cite{DBLP:journals/ijon/SuALPBL24} positional embeddings that are employed to train state-of-the-art models~\cite{dubey2024llama3herdmodel}.
To accelerate inference and reduce memory requirements, we employed grouped-query attention~\cite{DBLP:conf/emnlp/AinslieLJZLS23}. 
An entire overview of our architectural choices is presented in \Cref{table:model_hyper_parameters}.

These and other design decisions were guided by medium-scale (Chinchilla-optimal~\cite{DBLP:conf/nips/HoffmannBMBCRCH22} training of a 2.6B~parameter model) ablation runs for various training-related hyperparameters.
Our goal with these ablations was to find improvements in the compute-equivalent setting while also confirming whether proposed modifications transfer to our codebase, which is not necessarily the case~\cite{narang2021transformer}.
Appendix~\ref{sec:training-ablations-appendix} describes the conducted ablation experiments in detail, Table~\ref{tab:training-ablations} summarizes the results, and Figures~\ref{fig:training-ablations-plots} and~\ref{fig:training-ablations-plots-2} contain loss curves for the various experiments.

\subsection{Initial Training}\label{section:model_training}

Using the causal language modelling training objective, we trained our model on 6T tokens covering all 24 European languages as described in~\Cref{section:data}, employed AdamW as an optimizer and used a cosine learning rate schedule starting with a learning rate of 3e-5, increasing it to the maximum learning rate of 3e-4 within the first 10,000 steps, and decaying it afterwards.
During the training, we took two additional design decisions motivated by recent findings from related work.

\subsection{Adjusted Data Mixture}
\begin{figure*}[ht!]
    \centering
    \includegraphics[width=1\linewidth]{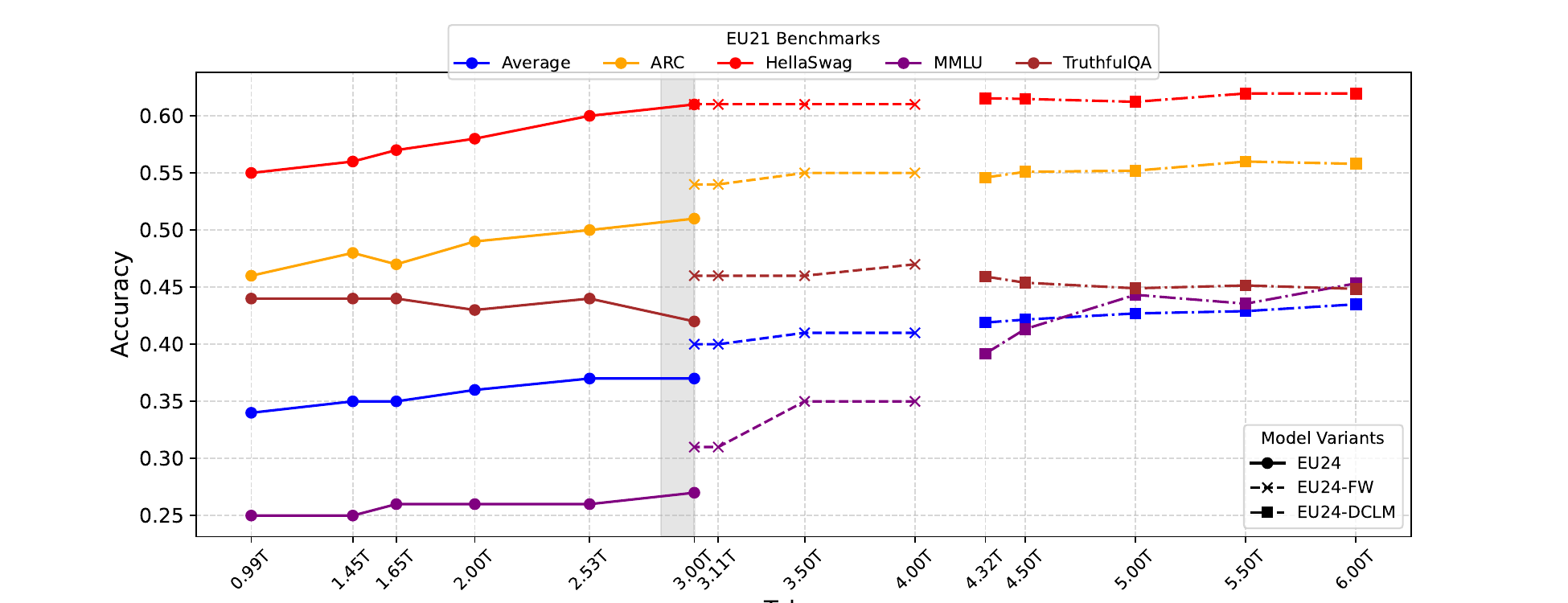}
    \caption{Downstream performance of the base model from 0.99T to 6T tokens across 21 European languages. The grey area highlights the ablation comparing the performance of EU24 data to the FineWeb-EDU dataset between 2.85T and 3T tokens. After 4T tokens, we replaced the English FineWeb-EDU and continued training until 6T tokens with DCLM-Baseline.}
    \label{fig:fine_web_edu_oscar}
\end{figure*}

Penedo \textit{et al.}~\cite{DBLP:journals/corr/abs-2406-17557} showed that training based on educational content further improves the performance of \acp{LLM}.
Therefore, after 2.85T tokens, we conducted an ablation where we trained one model up to 3T tokens based on FineWeb-EDU~\cite{DBLP:journals/corr/abs-2406-17557} (which is only available in English), and a second model with the initial data composition (see \Cref{section:data}).
As shown in \Cref{fig:fine_web_edu_oscar}, the model trained based on education content obtained an average performance improvement of 3\% across languages and benchmarks. 
In the appendix, we show the performance development for English (\Cref{fig:ducational_content_en}), French (\Cref{fig:ducational_content_fr}), German (\Cref{fig:ducational_content_de}), Finish (\Cref{fig:ducational_content_fi}) and Estonian (\Cref{fig:ducational_content_et}).

\begin{figure}[h!]
    \centering
    \includegraphics[width=\linewidth]{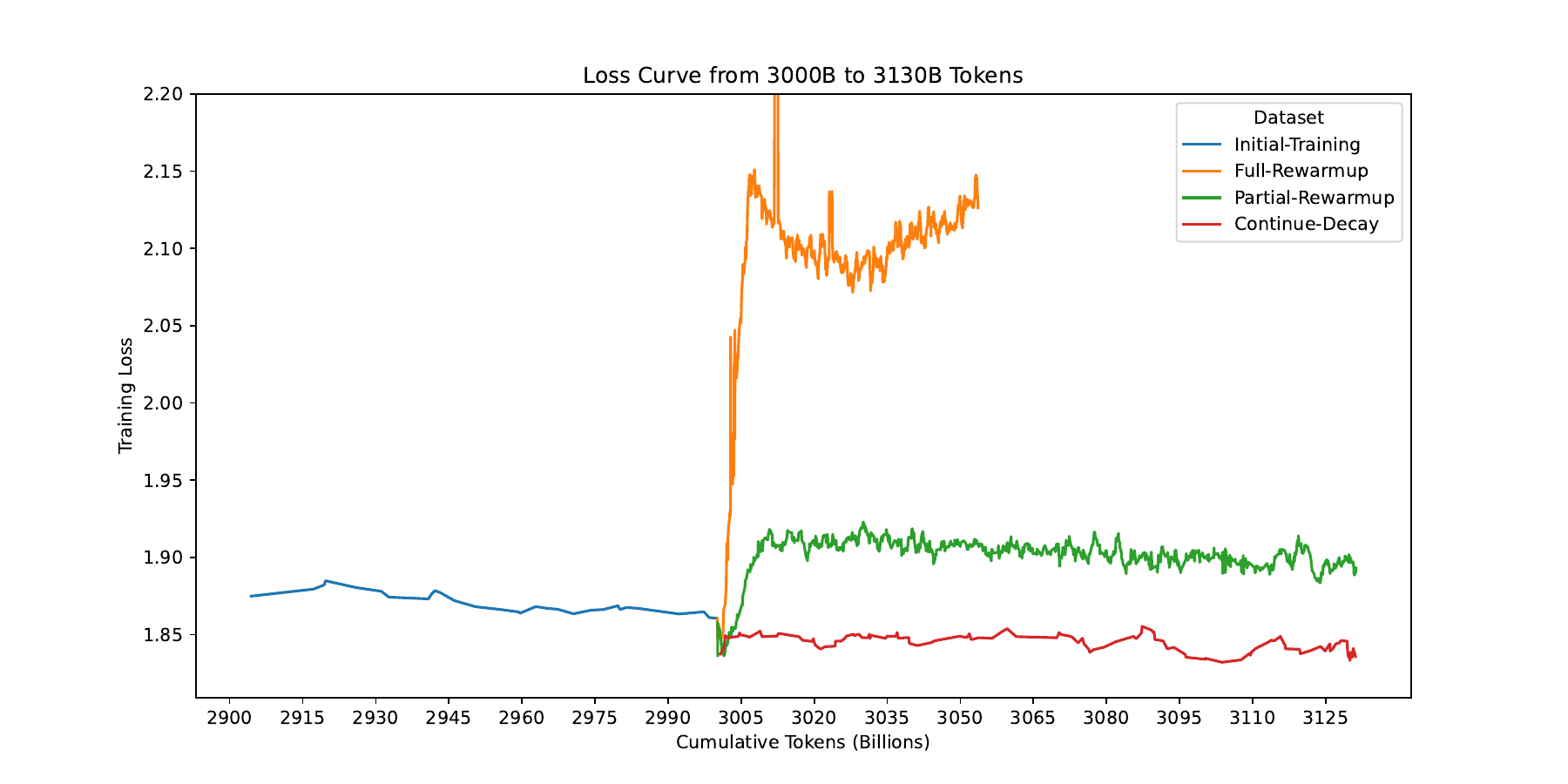}
    \caption{Loss curves up to 130B tokens for the investigated learning rate schedules.}
    \label{fig:learning_rate_schedules}
\end{figure}

\subsection{Continued Pre-Training}\label{appendix_continued_pre_training}
We want to emphasize that the initial goal has been to train our base model up to 3T tokens, which are beyond compute-optimal based on the scaling laws presented by Hoffmann \textit{et al.}~\cite{DBLP:conf/nips/HoffmannBMBCRCH22} for our model size. Recent works show that it is beneficial to further the pre-train model, e.g., LLaMA 3 has been trained up to 15T tokens, we decided to train the model on an additional 3T tokens and therefore needed to define a new learning rate schedule.
We ablated three different learning rate schedules \textit{Full-Rewarmup}, \textit{Partial-Rewarmup}, and \textit{Continue Decay}.
In the Full-Rewarmup setting, we defined a cosine learning rate schedule with the same minimal and maximal learning as for the initial pre-training phase.
In the Partial-Rewarmup, we set the maximum learning rate to a quarter of the maximum learning rate of the initial pre-training phase.
Finally, in the Continue-Decay setting, we decreased the learning rate to 5\% of the maximum learning rate within the following 3T tokens, providing the lowest training loss (\Cref{fig:learning_rate_schedules}).
Therefore, we decided to employ this setting to train our model further.

\section{Instruction Tuned Model}\label{section:instruction_tuned_model}

In the following, we describe our post-training procedure to instruction-tune Teuken-7B-Base resulting in Teuken-7B-Instruct, starting with a description of the respective fine-tuning stages (\Cref{sec:it_training}), followed by details about the used dataset composition (\Cref{sec:it_data}).

\subsection{Post-Training}\label{sec:it_training}

For instruction-tuning, we follow a common two-step approach with a first supervised finetuning (SFT) stage to enable the model to follow general instructions and a subsequent stage of preference optimization to refine the model's response behaviour.

\paragraph{Supervised finetuning (SFT).}
In SFT, models are provided with an instruction as an input and are optimized to produce a respective response as an output.
Our SFT setup follows common practice, with hyperparameter settings optimized via a small grid search (for details on hyperparameters we refer to Appendix~\ref{app:it_hyperparams}). 

In addition to the general setup, we employ noisy embeddings \cite{jain2023neftune} with $\alpha$ = 5; we mask the gradient of the input sequence and only update the learning signal from the response. Ultimately, we employ sequence packing to improve fine-tuning efficiency and optimize GPU usage~\cite{wang2024packing}.

\paragraph{Preference optimization.}
After the SFT stage, we further improve the model with direct-preference optimization (DPO)~\cite{rafailov2023direct}.
Again, we optimize hyperparameters on a small grid search (detail in Appendix~\ref{app:it_hyperparams}), and equivalent to SFT, we mask the input sequences.
    
\subsection{Data}
\label{sec:it_data}
Our dataset comprises publicly available datasets and a part that we synthesized. 
Inspired by the findings of \cite{DBLP:journals/corr/abs-2402-13703}, we utilized multilingual datasets to improve the cross-lingual performance of our model.
A list of datasets employed alongside their sample size and languages is presented in Table~\ref{tab:it_sft_data_composition_new} for SFT and \ref{tab:it_dpo_data_composition_new} for DPO, respectively.

\paragraph{Synthesized Data}
For generating SFT data, we follow the self-instruct paradigm \citep{wang2022self} with modified seed prompts and a different generation model \cite{DBLP:journals/corr/abs-2401-04088} ("Sigma").
To further enhance instruction diversity and complexity, we apply Evol-Instruct~\citep{xu2023wizardlm} using Mixtral, and regenerate the responses accordingly ("Sigma Evolved").
Additionally, we generate a small set of Everyday Conversations \citep{everydayconversations2024} using Mixtral (``Sigma Everyday conversations'').
Finally, we manually curate 113 self-awareness data points designed to inform the model about its own identity ("Teuken Self-Awareness SFT"). 
These are oversampled by a factor of 10 during training.

\paragraph{Translation}

We use Mixtral-8x22B-Instruct to generate high-quality translations of various source datasets into German, French, and Italian. 
Mixtral was selected after evaluating multiple LLMs of different scales on a representative subset of the SFT data, where it achieved the best translation quality among models with permissive licenses. 
Translation quality was assessed using Llama-3.1-70B-Instruct \citep{dubey2024llama3herdmodel} as an automated judge, showing strong agreement with an independent human annotator.

\paragraph{Data selection}

We use the deita quality and complexity scorer \citep{liu2023makes} to evaluate all instruction–response pairs. 
After normalizing scores across the full dataset, we compute a general preference score for each pair as a weighted sum of the two components\footnote{Weights used: $w_\text{quality} = 0.7$, $w_\text{complex} = 0.3$}. 
We then sample subsets of the source datasets by prioritizing data points with high preference scores.

For datasets we consider overly homogeneous, we further filter by selecting only data points that exhibit a minimal embedding distance (based on cosine similarity) to previously chosen samples (see Min Distance in Table~\ref{tab:it_sft_data_composition_new}). Embeddings are obtained by encoding all instructions using a SentenceTransformer model\footnote{\url{https://hf.co/sentence-transformers/distiluse-base-multilingual-cased-v1}} \citep{reimers-2019-sentence-bert,reimers-2020-multilingual-sentence-bert}.

\section{Results}

In the following, we present the results of our base and instruction-tuned models, with a particular focus on multilingual evaluation. 
Our evaluation targets the official European languages, as this aligns with the core objective of our research. 

\Cref{section:evaluation_set_up} describes our evaluation set-up, \Cref{section:evaluation_21_languages} presents our results across all 21 investigated languages, while \Cref{section:evaluation_6_languages} focuses on the performance of our models in the six widely spoken languages: English, German, French, Italian, Spanish, and Portuguese, and \Cref{section:evaluation_exclusive_languages} presents our findings on the remaining 15 languages. 
Additionally, \Cref{section:evaluation_common_languages} highlights the performance of various multilingual models on the largest common language subset.
Further results comparing our base and instruction-tuned model to related models across different language sets and individual languages are presented in the appendix (Table~\ref{tab:21_languages_base}–Table~\ref{tab:Swedish}).
Finally, \Cref{sec:it_eval} presents the instruction following capabilities of Teuken-7B-Instruct on the recently published multilingual benchmark MT-Bench-X~\cite{DBLP:journals/corr/abs-2402-13703}.

\subsection{Evaluation Set-Up}\label{section:evaluation_set_up}

\paragraph{\textbf{Evaluation Datasets}}The evaluation was conducted using ARC~\cite{DBLP:journals/corr/abs-1803-05457}/EU21-ARC~\cite{thellmann2024crosslingual} (25-shot for science-based questions), HellaSwag~\cite{DBLP:conf/acl/ZellersHBFC19}/EU21-HeSw~\cite{thellmann2024crosslingual} (10-shot for commonsense reasoning), TruthfullQA~\cite{DBLP:conf/acl/LinHE22}/EU21-TQA~\cite{thellmann2024crosslingual} (6-shot pseudo-shot for generating truthful answers), and MMLU~\cite{DBLP:conf/iclr/HendrycksBBZMSS21}/EU21-MMLU~\cite{thellmann2024crosslingual} (5-shot for broad knowledge), which are available in 21 of the 24 official European languages, for which we report mean accuracy.

\paragraph{\textbf{Models}} We evaluated our models against related multilingual models with up to 8B parameters that have been pre-trained on causal language modelling from scratch and models that result from an instruction/fine-tuning of the pre-trained model. 
We did not include models that represent the distillation of larger models to ensure a comparable setting.
As a result, we evaluated against Aya-23 8B, Bloom-7B1, Bloomz-7B1, Meta-Llama-3.1-8B, Meta-Llama-3.1-8B Chat, Salamandra-7B, Salamandra-7B Instruct,  Pharia-1-LLM-7B,  Pharia-1-LLM-7B-C-A (Control-Aligned), Occiglot-7B-EU5, and Mistral. 
\Cref{table:base_models} provides an overview of the number of languages and tokens that the models have been trained on. 
Note that the Mistral models were not trained as multilingual models and are intended to be used as English-only models.
However, they contain multilingual to some extent, as described in~\cite{DBLP:journals/corr/abs-2401-04088}.
Due to its strong performance, we included Mistral in our evaluations.

\begin{table}[h]
\centering
\begin{tabular}{lcc}
\toprule
\textbf{Model} & \textbf{Languages} & \textbf{Tokens} \\ 
\midrule
Aya-23 8B & 23 & N/A \\ 
Bloom-7B1 & 46 & 350B \\ 
Bloomz-7B1 & 46 & 354B \\ 
Meta-Llama-3.1-8B & 8 & 15T \\
Mistral-7B-v0.3 & 1 & N/A \\ 
Salamandra-7B & 35 & 7.8T \\ 
Pharia-1-LLM-7B & 7 & 4.7T \\ 
\midrule
Teuken-7B-Base (Ours) & 24 & 6T \\ 
\bottomrule
\end{tabular}
\caption{Evaluated models and the number of languages and training tokens the models have been trained on.}
\label{table:base_models} 
\end{table}

\subsection{Performance on 21 European Languages}\label{section:evaluation_21_languages}

\Cref{tab:21_languages_instruct} and \Cref{tab:21_languages_base} present the multilingual results of the instruction-tuned and based models for all 21 European languages.
\Cref{tab:21_languages_instruct} illustrates that our instruction-tuned model leads with an average accuracy of 57.0\% across all benchmarks, followed by Meta-Llama-3.1-8B-Instruct. 
The results are remarkable, considering our model has been trained on significantly fewer tokens than Llama-3.1-8B-Instruct (see Table \ref{table:base_models}). 
We hypothesize that the training on the FineWeb-EDU dataset has contributed significantly to this aspect since it has shown to provide similar results compared to different composed datasets while requiring significantly fewer tokens~\cite{DBLP:journals/corr/abs-2406-17557}.
Noteworthy, it improves cross-lingual performance (see \Cref{fig:fine_web_edu_oscar}) even though the dataset is in English.
On benchmark-level, Meta-Llama-3.1-8B-Instruct excels in EU21-MMLU (57.6\%) while Salamandra-7B-Instruct is particularly strong in EU21-ARC (59.5\%).

\begin{figure*}
    \centering
    \includegraphics[width=0.9\linewidth]{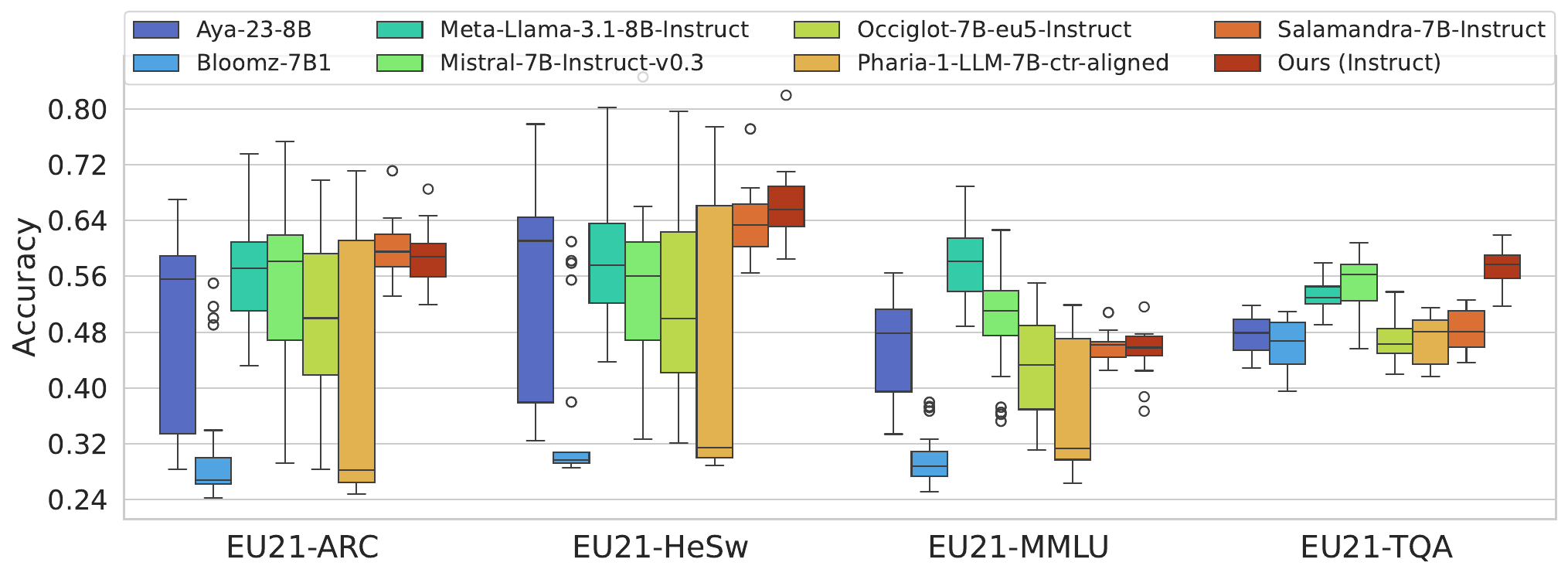}
    \caption{Models' performance across 21 languages for different benchmarks.}
    \label{fig:boxplots_21_languages}
\end{figure*}

Besides the average performance across languages, we also investigated the robustness across languages. 
\Cref{fig:boxplots_21_languages} shows that our model and Salamandra-7B-Instruct and Bloomz-7B1 are significantly more robust than the other models on all benchmarks except for EU21-TQA where several models obtain robust performance across languages, ensuring that the model performs comparably across languages.

\subsection{Performance on Top-6 European Languages}\label{section:evaluation_6_languages}

\Cref{tab:6_languages_instruct} focuses on the performance of the models on six widely spoken European languages: English, German, French, Italian, Spanish, and Portuguese. 
These languages are well-represented in training data and commonly used to evaluate multilingual models.
Meta-Llama-3.1-8B-Instruct leads with 62.3\% average performance across tasks, excelling in EU21-MMLU (63.2\%), followed closely by Mistral-7B-Instruct-v0.3 with 61.3\% on average, obtaining the best performance in EU21-TQA (56.8\%) and EU21-ARC (65.4\%). 
Teuken-7B-Instruct (Ours) outperforms all models on EU21-HeSw with 71.9\% and performs competitively in all remaining benchmarks except for EU21-MMLU.

\subsection{Performance on Exclusive European Languages}\label{section:evaluation_exclusive_languages}

\Cref{tab:15_languages_instruct} compares the models across 15 less commonly evaluated European languages (Romanian, Czech, Danish, Greek, Estonian, Finnish, Hungarian, Lithuanian, Latvian, Dutch, Bulgarian, Polish, Slovak, Slovenian, and Swedish).
Teuken-7B-Instruct (Ours) leads with an average accuracy of 55.9\%, excelling in EU21-HeSw (64.0\%) and EU21-TQA (58.3\%), followed by Meta-Llama-3.1-8B-Instruct that obtained the best performance on EU21-MMLU (55.4\%) and Salamandra-7B-Instruct that obtained the strongest performance on on EU21-ARC (57.6\%).
\begin{table*}

\begin{subtable}{\textwidth}
    \centering
        \begin{tabular}{lccccc}
            \toprule
            Model & Avg. & EU21-ARC & EU21-HeSw & EU21-TQA & EU21-MMLU \\ 
            \midrule
            Meta-Llama-3.1-8B-Instruct & \underline{.563} & .563 & .579  & .532  & .\textbf{576}  \\
            Mistral-7B-Instruct-v0.3 & .527 & .530 & .538  & \underline{.548}  & \underline{.491}  \\
            Salamandra-7B-Instruct & .543 & \textbf{.595}  & \underline{.637} & .482  & .459 \\
            Aya-23-8B & .485 & .475 & .535  & .476 & .455  \\
            Occiglot-7B-eu5-Instruct & .475 & .484 & .519 & .471  & .428  \\
            Pharia-1-LLM-7B-C-A & .417 & .396 & .438  & .469 & .366\\
            Bloomz-7B1 & .358 & .316 & .354 & .461  & .302  \\
            \midrule
            Teuken-7B-Base (Ours) & .520 & .558  & .619  & .449 & .453\\
            Teuken-7B-Instruct (Ours) & \textbf{.570} & \underline{.590}  & \textbf{.663}  & \textbf{.573} & .454 \\
            \bottomrule
        \end{tabular}
        \caption{Results on multilingual benchmarks for 21 European languages with instruction-tuned models.}
        \label{tab:21_languages_instruct}
        \vspace{1cm}
\end{subtable}

\centering\begin{subtable}{\textwidth}
        \centering
        \begin{tabular}{lccccc}
            \toprule
            Model & Avg. & EU21-ARC & EU21-HeSw & EU21-TQA & EU21-MMLU \\ 
            \midrule
            Meta-Llama-3.1-8B-Instruct & \textbf{.623} & \underline{.648}  & .677 & .535  & \textbf{.632} \\
            Mistral-7B-Instruct-v0.3 & \underline{.613} & \textbf{.654}  & .670  & \textbf{.568 } & \underline{.560} \\
            Aya-23-8B & .574 & .614  & .687  & .470  & .526  \\
            Occiglot-7B-eu5-Instruct & .583 & .646  & \underline{.712}  & .458  & .518  \\
            Pharia-1-LLM-7B-C-A & .580 & .643  & .696  & .497  & .485  \\
            Salamandra-7B-Instruct & .565 & .643  & .685 & .455  & .477  \\
            Bloomz-7B1 & .433 & .449  & .502  & .426 & .354  \\
            \midrule
            Teuken-7B-Base (Ours) & .541 & .608 & .674  & .405  & .478  \\
            Teuken-7B-Instruct (Ours) & .598 & .641 & \textbf{.719}  & \underline{.549}  & .482  \\
            \bottomrule
        \end{tabular}
        \caption{Results on multilingual benchmarks for 6 Languages (English, German, French, Italian, Spanish, Portuguese) for instruction-tuned models.}
        \label{tab:6_languages_instruct}
        \vspace{1cm}

\end{subtable}

\begin{subtable}{\textwidth}
        \centering
        \begin{tabular}{lccccc}
            \toprule
            Model & Avg. & EU21-ARC & EU21-HeSw & EU21-TQA & EU21-MMLU \\ 
            \midrule
            Meta-Llama-3.1-8B-Instruct & \underline{.538} & .529  & .540  & .530  & \textbf{.554}  \\
            Mistral-7B-Instruct-v0.3 & .492 & .480  & .485  & \underline{.540}  & \underline{.463}  \\
            Salamandra-7B-Instruct & .535 & \textbf{.576} & \underline{.619}  & .493 & .451 \\
            Aya-23-8B & .450 & .420 & .473 & .479 & .427 \\
            Occiglot-7B-eu5-Instruct & .432 & .419 & .441  & .476 & .392  \\
            Pharia-1-LLM-7B-C-A & .352 & .296 & .334 & .458 & .319  \\
            Bloomz-7B1 & .328 & .263 & .295 & .475 & .281 \\
            \midrule
            Teuken-7B-Base (Ours) & .511 & .539 & .597 & .466 & .443 \\
            Teuken-7B-Instruct (Ours) & \textbf{.559} & \underline{.569} & \textbf{.640} & \textbf{.583} & .443   \\
            \bottomrule
        \end{tabular}
        \caption{Results on multilingual benchmarks across the 15 exclusive European languages (Romanian, Czech, Danish, Greek, Estonian, Finnish, Hungarian, Lithuanian, Latvian, Dutch, Bulgarian, Polish, Slovak, Slovenian, and Swedish) for instruction-tuned models.}
        \label{tab:15_languages_instruct}
        \vspace{1cm}

\end{subtable}

\vspace{1em}

\begin{subtable}{\textwidth}
        \centering
        \begin{tabular}{lccccc}
            \toprule
            Model & Average & EU21-ARC & EU21-HeSw & EU21-TQA & EU21-MMLU \\ 
            \midrule
            Aya-23-8B & \underline{.563} & .593  & .661  & \underline{.484}  & \textbf{.512}  \\
            Salamandra-7B-Instruct & .557 & \underline{.620}  & .664  & .476  & \underline{.468} \\
            Bloomz-7B1 & .394 & .373  & .419  & .456  & .327  \\
            \midrule
            Teuken-7B-Instruct (Ours) & \textbf{.585} & \underline{.619} & \textbf{.692}  & \textbf{.564}  & .464  \\
            \bottomrule
        \end{tabular}
        \caption{Instruct model results on multilingual benchmarks across the 10 common languages (Czech, Dutch, English, French, Greek, Italian, Polish, Portuguese, Romanian, and Spanish). The tables include mean accuracy across languages.}
        \label{tab:11_languages_instruct}
        \end{subtable}
\end{table*}

\subsection{Comparison on Common Languages}\label{section:evaluation_common_languages}

\Cref{tab:11_languages_instruct} compares the performance of Teuken-7B-Instruct (Ours) models with Aya-23-8B, Bloomz-7B1, and Salamandra-7B-Instruct across ten common European languages: Czech, Dutch, English, French, Greek, Italian, Polish, Portuguese, Romanian, and Spanish.
Teuken-7B-Instruct (Ours) leads with an average of 58.5\%, obtaining the best results in EU21-HeSw (69.2\%) and EU21-TQA (56.4\%). 
Aya-23-8 B's, the second best-performing model, achieves an average performance of 56.3\%, excelling on EU21-MMLU (51.2\%).

Overall, the results highlight the strong performance of our model across various language sets.

\subsection{Multilingual Instruction Following Evaluation}\label{sec:it_eval}

For evaluating the multilingual instruction-following capabilities of Teuken-7B-Instruct, we utilized MT-Bench-X~\cite{DBLP:journals/corr/abs-2402-13703} that employs an LLM as a judge across all five available evaluation languages: English, German, French, Italian, and Spanish. 
The results are presented in the \Cref{apx:it_eval_results} and \Cref{fig:radar-gpt-4-mt-bench-X}.
We compare the cross-lingual performance of our model with Llama-3.1-8B-Instruct, Mistral-7B-Instruct-v0.3 and Salamandra-7B-Instruct in the following.
Overall, Teuken-7B-Instruct demonstrates robust cross-lingual performance in several key domains.
In particular, the Teuken-7B-Instruct model exhibits strengths in creative tasks such as Writing and Roleplay and in knowledge-based domains like Humanities and Stem, especially in German. However, it noticeably underperforms in the Math and Coding categories, highlighting areas for potential future optimization.
This performance gap can be explained by the fact that our model has not been optimized for these capabilities.
Compared to Salamandra-7B-Instruct, our model provides superior multilingual capabilities across most tasks and languages. Notable exceptions include the Reasoning tasks in Italian and German and the extraction task, where Salamandra-7B-Instruct achieves slightly better cross-lingual performance.

\subsubsection{Toxicity}

To measure the toxicity of our model, we compare our models against other instruction-tuned baselines using the PolygloToxicityPrompts benchmark \cite{jain2024polyglotoxicityprompts}, reporting results for the $\textsc{PTP}_{\text{SMALL}}$ subset.
Our evaluation setup follows the protocol described in \cite{jain2024polyglotoxicityprompts}, with one exception: we set the repetition parameter to $K=1$.
In addition to the benchmark's core metrics, we report toxicity and profanity scores using the Perspective API \cite{leesperspectiveapi}.
For each attribute $i \in {\text{Profanity}, \text{Toxicity}}$, we compute both the Empirical Probability ($\text{EP}_i$) and the Average Score ($\text{A}_i$).
\Cref{tab:ptp_agg} presents the aggregated results across five languages: German, English, French, Italian, and Spanish. Detailed language-specific scores are provided in \Cref{tab:ptp_english}–\ref{tab:ptp_spanish} in the appendix.

\begin{table}[H]
    \centering
    \begin{tabular}{lcccc}
        \toprule
        Model & $\textsc{EP}_{\textit{Prof.}}$ & $\textsc{A}_{\textit{Prof.}}$ & $\textsc{EP}_{\textit{Tox.}}$ & $\textsc{A}_{\textit{Tox.}}$ \\
        \midrule
        Meta-Llama-3.1-8B-Instruct & .174 & .196 & .173 & .219 \\
        Mistral-7B-Instruct-v0.3 & .139 & .160 & .140 & .183 \\
        Salamandra-7B-Instruct & .206 & .216 & .196 & .230 \\
        Aya-23-8B & .177 & .197 & .176 & .218 \\
        Occiglot-7B-eu5-Instruct & .202 & .212 & .193 & .226 \\
        Bloomz-7B1 & .100 & \textbf{.115} & .106 & \textbf{.134} \\
        \midrule
        Teuken-7B-Instruct (Ours) & \textbf{.081} & .149 & .076 & \textbf{.152} \\
        \bottomrule
    \end{tabular}
    \caption{The aggregated $\textsc{PTP}_{\text{small}}^{\text{Aggregated}}$ evaluated on instruction-tuned models.}
    \label{tab:ptp_agg}
\end{table}

Among the evaluated models, Teuken-7B-Instruct (Ours) demonstrates superior per-
formance in minimizing the likelihood of generating harmful content. It achieves the lowest
EPProfanity of 0.081 and EPToxicity of 0.076, indicating that it is the least likely to produce pro-
fane or toxic responses. In comparison, Bloomz-7B1, the next best performer, has EPProfanity
of 0.100 and EPToxicity of 0.106. Other models, such as Salamandra-7B-Instruct and Occiglot-
7B-eu5-Instruct, exhibit significantly higher probabilities, with EPToxicity values exceeding 0.19,
The Teuken-7B-Instruct model excels in minimizing harmful content, with the lowest empirical probabilities for profanity (0.081) and toxicity (0.076) on the PolygloToxicityPrompts benchmark. This makes it less likely to generate harmful outputs compared to other models. Although Bloomz-7B1 has slightly lower severity scores when harmful content occurs (0.115 for profanity, 0.134 for toxicity) versus Teuken’s (0.149 and 0.152), Teuken’s lower occurrence rate results in a safer expected toxicity (0.0116) than Bloomz’s (0.0142). Other models, like Salamandra-7B-Instruct and Occiglot-7B-eu5-Instruct, perform worse, with expected toxicity values above 0.04. Thus, Teuken-7B-Instruct is a top choice for safety-focused applications.
highlighting Teuken’s substantial improvement in this regard.  In conclusion, Teuken-7B-Instruct stands out for its ability to reduce the likelihood of harmful outputs while maintaining competitive severity levels, making it a strong candidate for appli-
cations prioritizing safety and reliability.

\section{Conclusion \& Future Work}

In this work, we presented the development of two multilingual large language models, Teuken-7B-Base and Teuken-7B-Instruct, tailored to support the linguistic diversity of Europe by encompassing all 24 official EU languages. 
Through the use of a custom multilingual tokenizer and a dataset prioritizing non-English content, our models address limitations found in existing multilingual models, particularly their English-centric bias. 
Our results demonstrate strong performance across multiple benchmarks, including ARC, HellaSwag, MMLU, and TruthfulQA, while training on significantly fewer tokens compared to related models.
The presented multilingual models, focusing on European languages, are an excellent base for further pre-training or fine-tuning each European language. 
We have taken great care to ensure that all languages are represented in the tokenizer, promoting the inclusivity and applicability of our models in diverse linguistic contexts.
We shared insights regarding our model development, often not described in detail, to support future development aimed at developing multilingual models.

In the future, we aim to expand our efforts along four key dimensions.
First, we plan to enhance our models’ capabilities further, focusing on mathematics and coding. To achieve this, we will continue training and fine-tuning our models, placing special emphasis on filtering high-quality reasoning, math and code data from the web—a strategy shown to be effective~\cite{shao2024deepseekmath,gunasekar2023textbooks}. 
Additionally, we will synthesize new training examples to boost performance in these domains.
Second, we intend to curate even higher-quality pre-training datasets by leveraging (large) language models as judges in the filtering process, building on recent advancements~\cite{su2024nemotron}.
Third, we aim to broaden our linguistic coverage by training on a more diverse set of European languages beyond the 24 official EU languages, making our technology accessible to an even wider audience.
Finally, we plan to increase the size of the model to excel in increasingly complex tasks and push the performance boundaries.

\section*{Acknowledgments}

This work was funded by the German Federal Ministry for Economic Affairs and Climate Action (BMWK) through the project OpenGPT-X (project no. 68GX21007D) as well as by the Federal Ministry of Education and Research of Germany and the state of North-Rhine Westphalia as part of the Lamarr-Institute for Machine Learning and Artificial Intelligence, LAMARR22B, and by the European Union’s Horizon 2020 research and innovation program under grant agreement No 101135671 (TrustLLM).
The authors gratefully acknowledge the Gauss Centre for Supercomputing e.V. (www.gauss-centre.eu) for funding this project by providing computing time on the GCS Supercomputer JUWELS at Jülich Supercomputing Centre (JSC) as well as the Center for Information Services and High Performance Computing [Zentrum für Informationsdienste und Hochleistungsrechnen (ZIH)] at TU Dresden for providing its facilities for automatic evaluation computations.

\bibliography{mybibfile}
\onecolumn 
\appendix
\section{Appendix}
\label{sec:appendix}

\subsection{Data}\label{sec:dumps}
In this section, we outline the Common Crawl data used, detailing the cutoff dates and the data collection period. 
The dumps span multiple years, with varying distributions of weeks per year (cf.~\Cref{tab:dumps_by_year}), which is critical for understanding the temporal coverage of the training data.

\begin{table}
\centering
\begin{tabular}{cl}
\toprule
\textbf{Year} & \textbf{Week} \\ 
\midrule
2014         & 42               \\ 
2015         & 14, 48           \\ 
2016         & 22, 44           \\ 
2017         & 13, 47, 51       \\ 
2018         & 5, 9, 13, 17, 22, 26, 30, 34, 39, 43, 47, 51 \\ 
2019         & 4, 9, 13, 18, 22, 26, 30, 35, 39, 47, 51 \\ 
2020         & 5, 10, 16, 24, 29, 34, 40, 45, 50 \\ 
2021         & 4, 10, 17, 21, 25, 31, 39, 43, 49 \\ 
2022         & 5, 21, 27, 33, 40, 49 \\ 
2023         & 6, 14, 23, 40, 50 \\ 
\bottomrule
\end{tabular}
\caption{List of dumps by year and week.}
\label{tab:dumps_by_year}
\end{table}

The data spans from 2014 to 2023, with earlier years (2014-2016) containing fewer dumps but often representing a larger dataset per dump. 
The latest available data cutoff is from 2023 (week 50), giving the model a comprehensive span of nearly a decade of web data.

This periodization ensures that the model has exposure to a wide temporal range of web content, with a balanced emphasis on both earlier, higher-density dumps and more recent, frequent dumps, providing a robust training corpus.

\subsection{Training Ablation Experiments}\label{sec:training-ablations-appendix}

\begin{figure*}[t]
    \centering
    
    \begin{subfigure}[t]{0.49\linewidth}
        \centering
        \includegraphics[width=\linewidth]{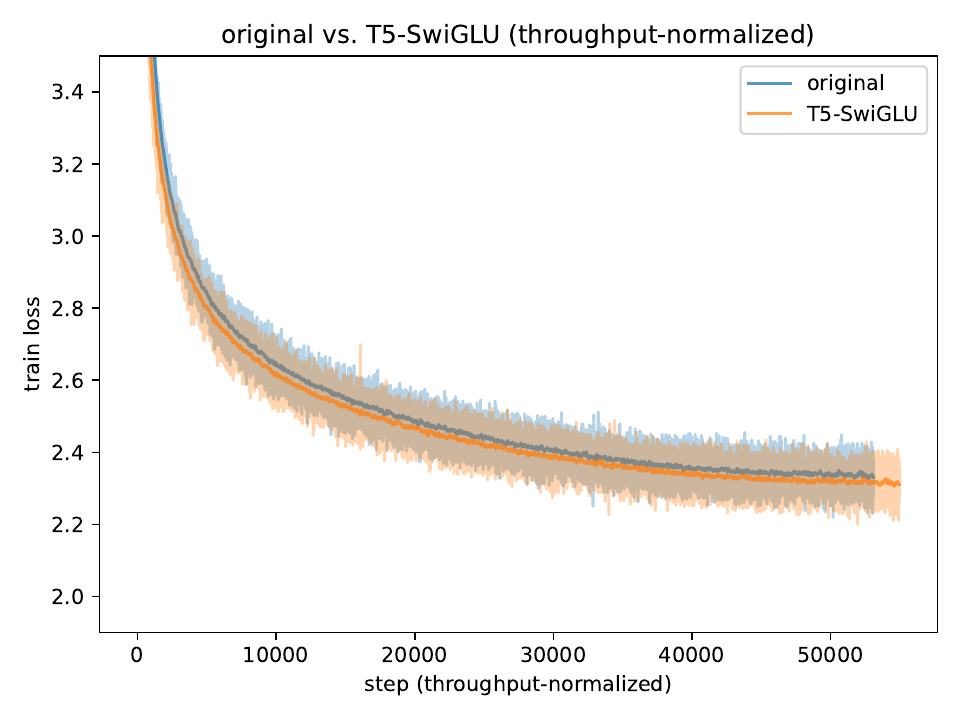}
        \caption{``SwiGLU'' ablation (time-normalized)}\label{fig:training-swiglu}
    \end{subfigure}
    \hfill
    \begin{subfigure}[t]{0.49\linewidth}
        \centering
        \includegraphics[width=\linewidth]{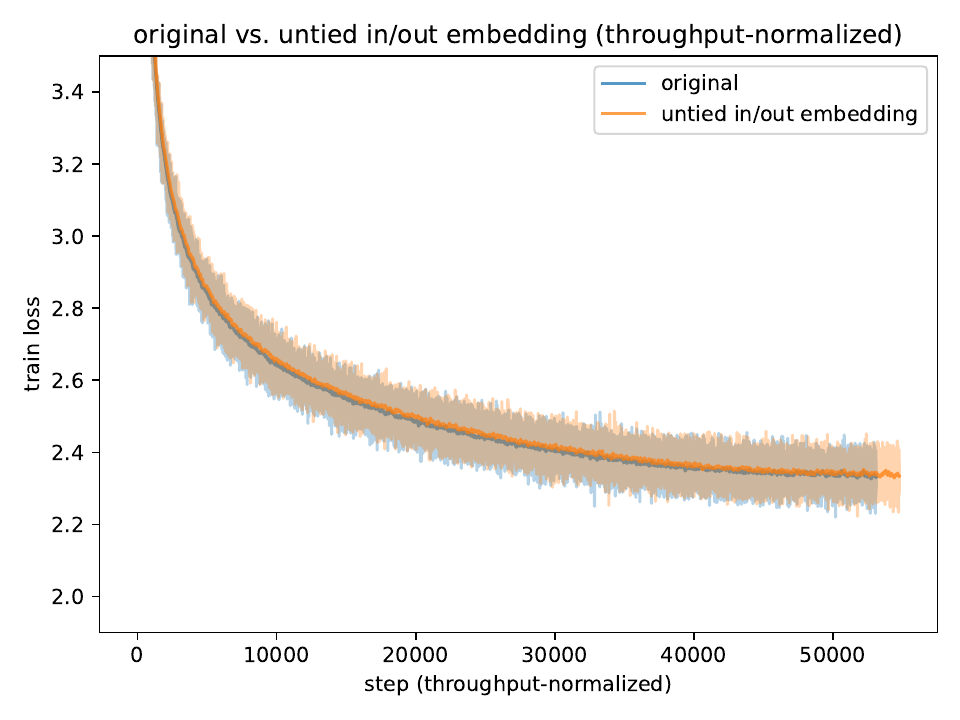}
        \caption{``Untied in/out embedding'' ablation (time-normalized)}\label{fig:training-untied-in-out-embedding}
    \end{subfigure}
    
    \vspace*{0.2cm} %
    
    \begin{subfigure}[t]{0.49\linewidth}
        \centering
        \includegraphics[width=\linewidth]{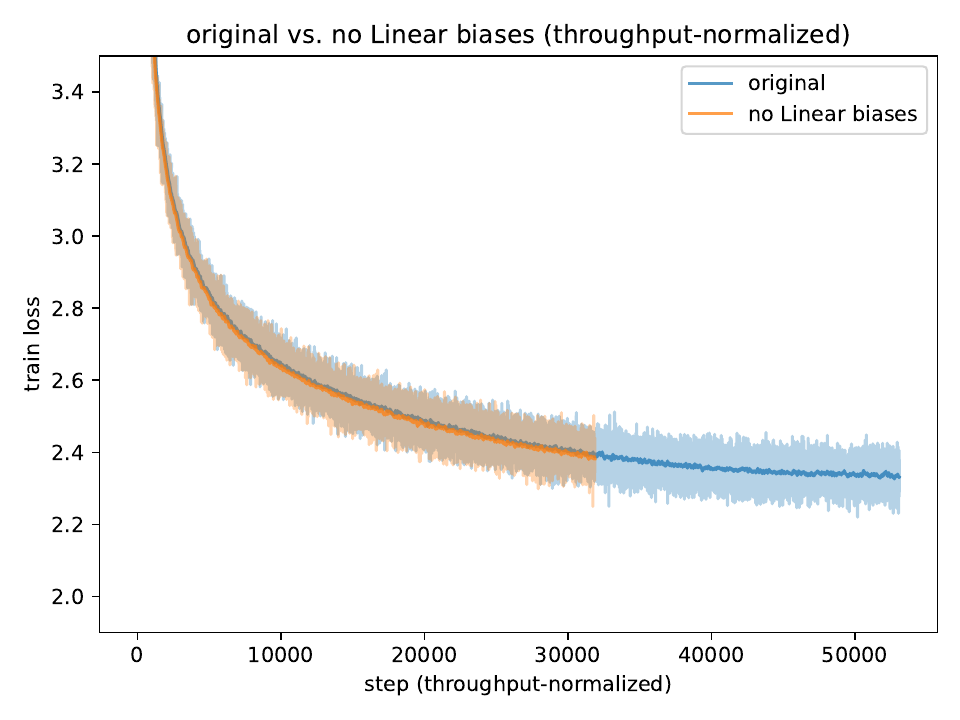}
        \caption{``No \texttt{Linear} biases'' ablation (time-normalized)}\label{fig:training-no-linear-biases}
    \end{subfigure}
    \hfill
    \begin{subfigure}[t]{0.49\linewidth}
        \centering
        \includegraphics[width=\linewidth]{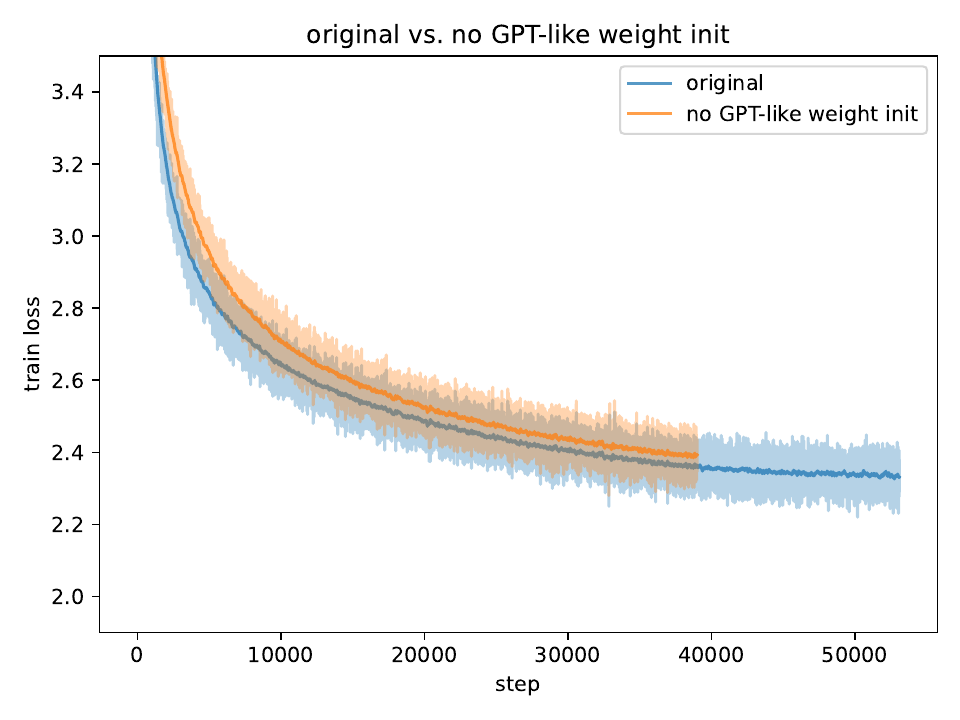}
        \caption{``No GPT-like weight init'' ablation}\label{fig:training-no-gpt-like-weight-init}
    \end{subfigure}
    
    \vspace*{0.2cm} %
    
    \begin{subfigure}[t]{0.49\linewidth}
        \centering
        \includegraphics[width=\linewidth]{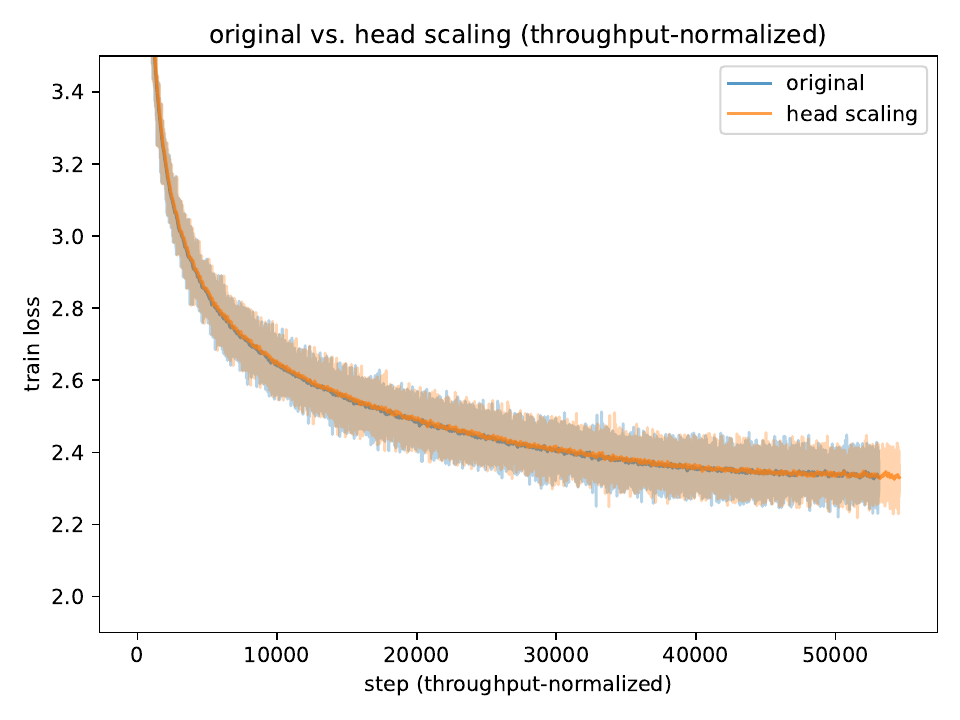}
        \caption{``Head scaling'' ablation (time-normalized)}\label{fig:training-head-scaling}
    \end{subfigure}
    \hfill
    \begin{subfigure}[t]{0.49\linewidth}
        \centering
        \includegraphics[width=\linewidth]{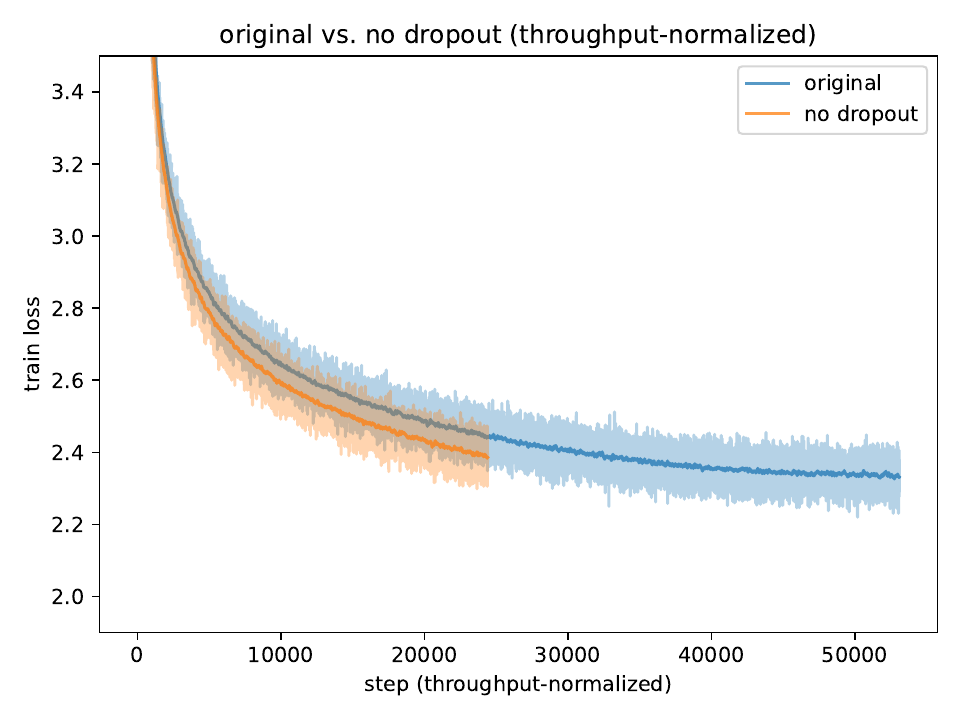}
        \caption{``No dropout'' ablation}\label{fig:training-no-dropout}
    \end{subfigure}
        \vspace*{0.3cm} %

    \caption{Training loss curves using (a)~SwiGLU, (b)~untied input and output embeddings, (c)~no \texttt{Linear} biases, (d)~no GPT-like weight initialization, (e)~per-head scaling factors, and (f)~no dropout. All plotted against the baseline.}\label{fig:training-ablations-plots}
\end{figure*}

\begin{figure*}[t]
    \centering
    \begin{minipage}{\linewidth}
        \centering
        \begin{minipage}{0.49\linewidth}
            \centering
            \includegraphics[width=\linewidth]{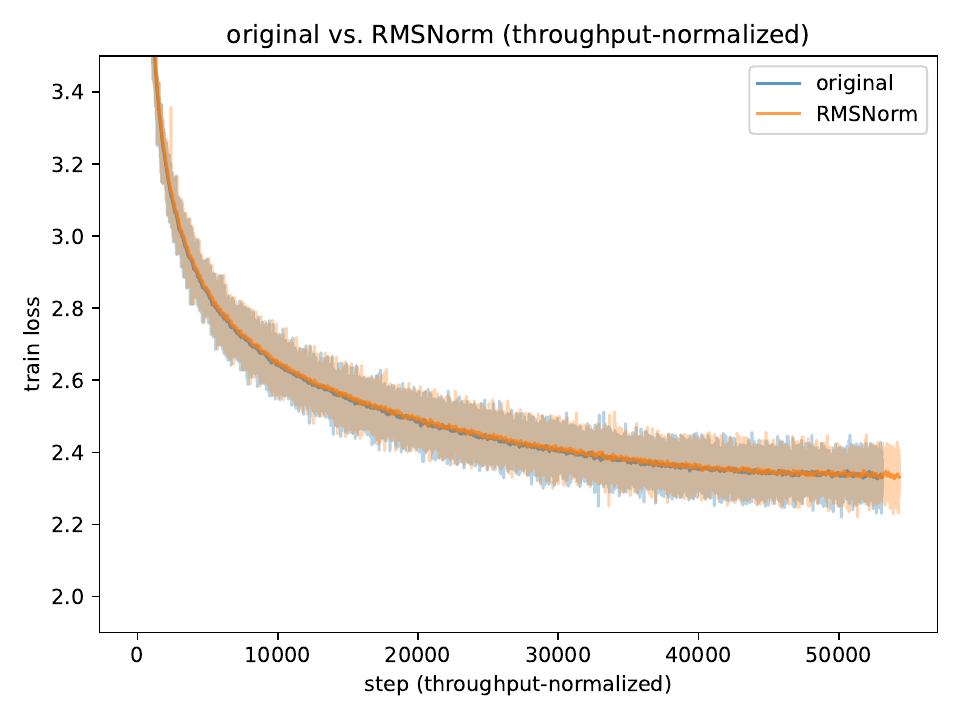}
            \captionof{figure}{``RMSNorm'' ablation (time-normalized)}\label{fig:training-rmsnorm}
        \end{minipage}
        \hfill
        \begin{minipage}{0.49\linewidth}
            \centering
            \includegraphics[width=\linewidth]{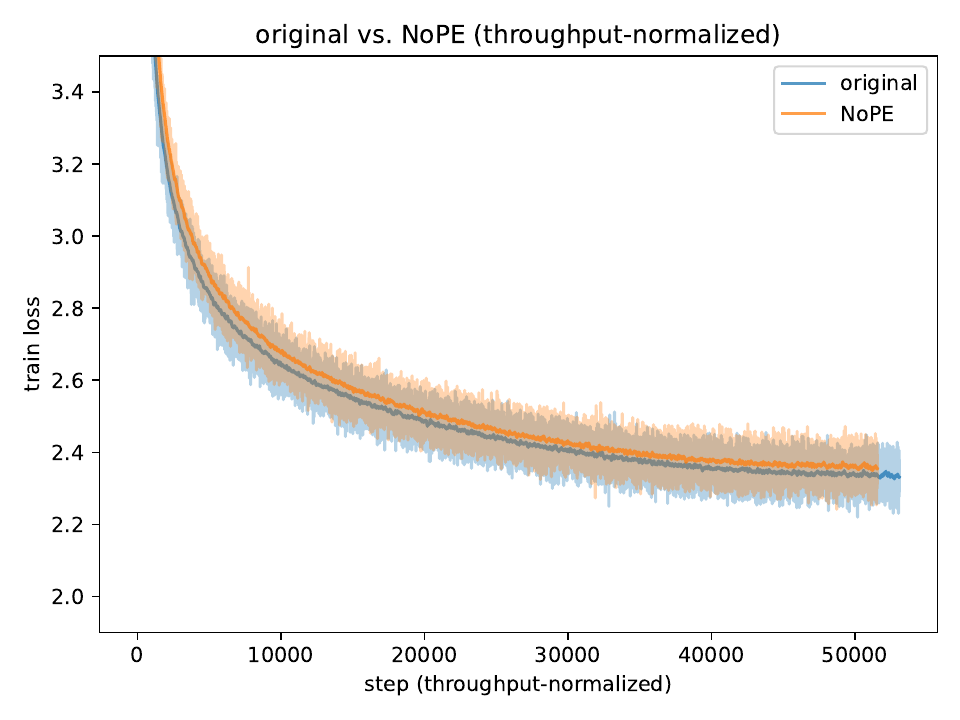}
            \captionof{figure}{``NoPE'' ablation (time-normalized)}\label{fig:training-nope}
        \end{minipage}
    \end{minipage}
    
    \vspace*{0.2cm}
    
    \begin{minipage}{\linewidth}
        \centering
        \begin{minipage}{0.49\linewidth}
            \centering
            \includegraphics[width=\linewidth]{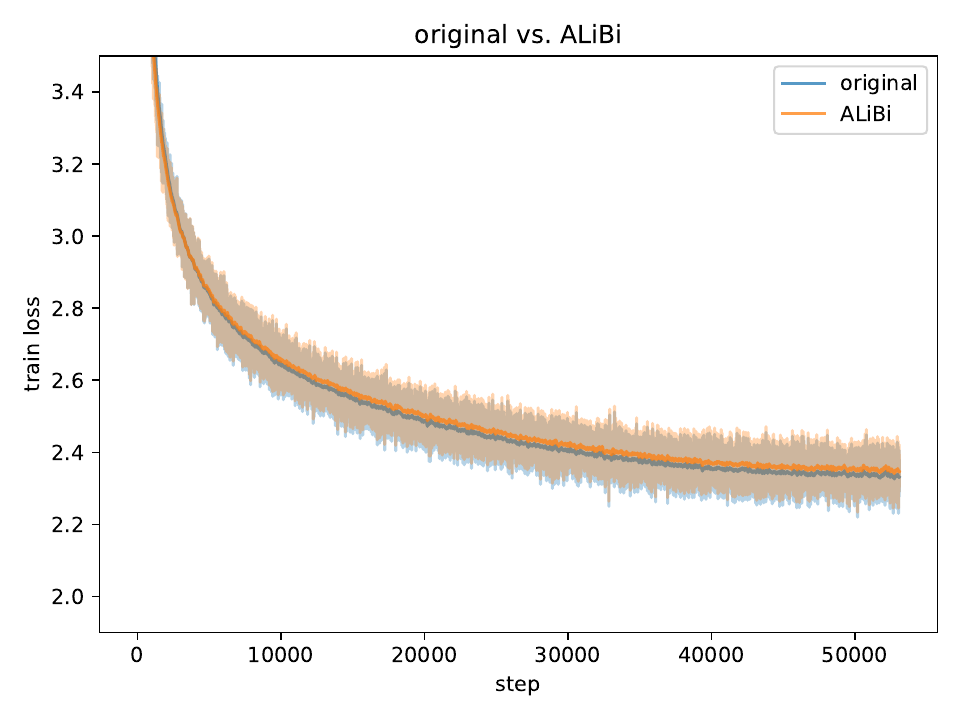}
            \captionof{figure}{``ALiBi'' ablation (\emph{not} time-normalized out of fairness)}\label{fig:training-alibi}
        \end{minipage}
        \hfill
        \begin{minipage}{0.49\linewidth}
            \centering
            \includegraphics[width=\linewidth]{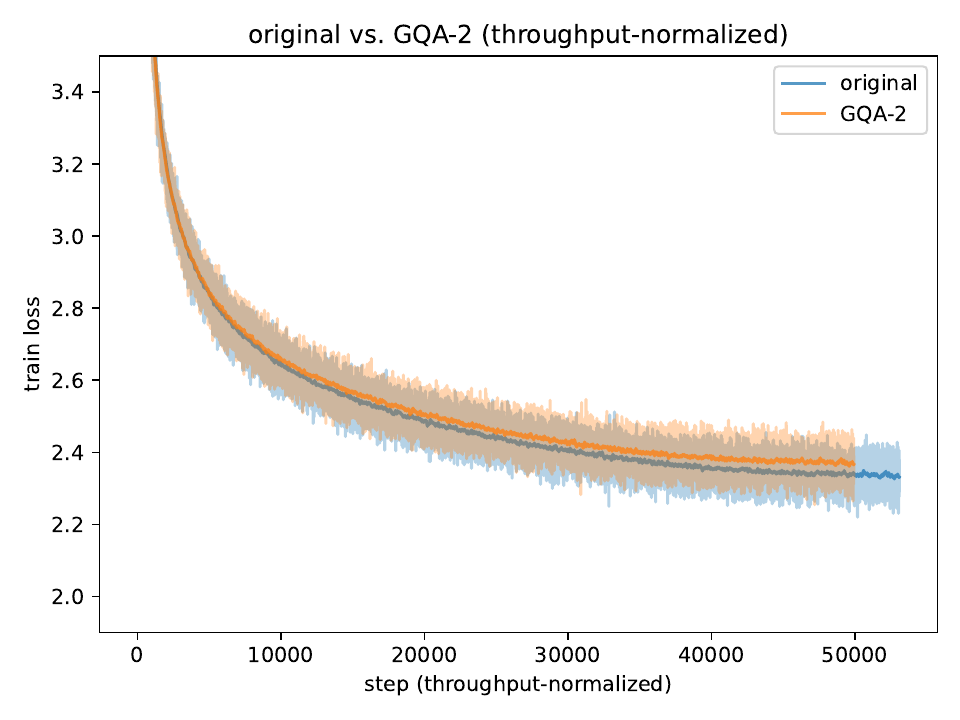}
            \captionof{figure}{``GQA~(2~groups)'' ablation (time-normalized)}\label{fig:training-gqa-2-groups}
        \end{minipage}
    \end{minipage}
    
    \vspace*{0.2cm}
    
    \begin{minipage}{\linewidth}
        \centering
        \begin{minipage}{0.49\linewidth}
            \centering
            \includegraphics[width=\linewidth]{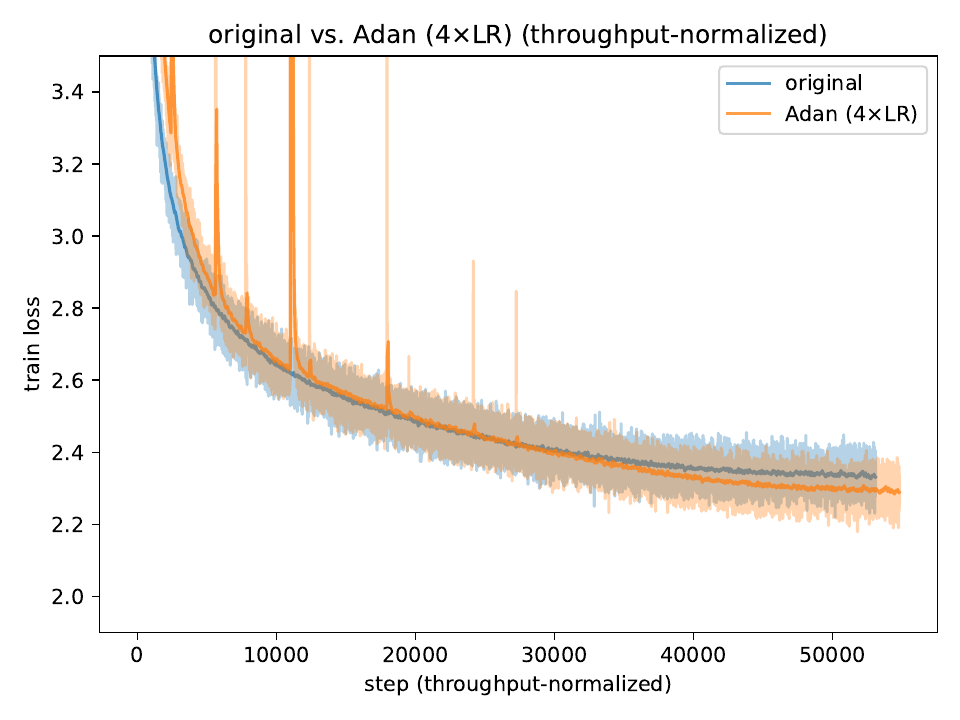}
            \captionof{figure}{``Adan'' ablation (time-normalized)}\label{fig:training-adan}
        \end{minipage}
        \hfill
        \begin{minipage}{0.49\linewidth}
            \centering
            \includegraphics[width=\linewidth]{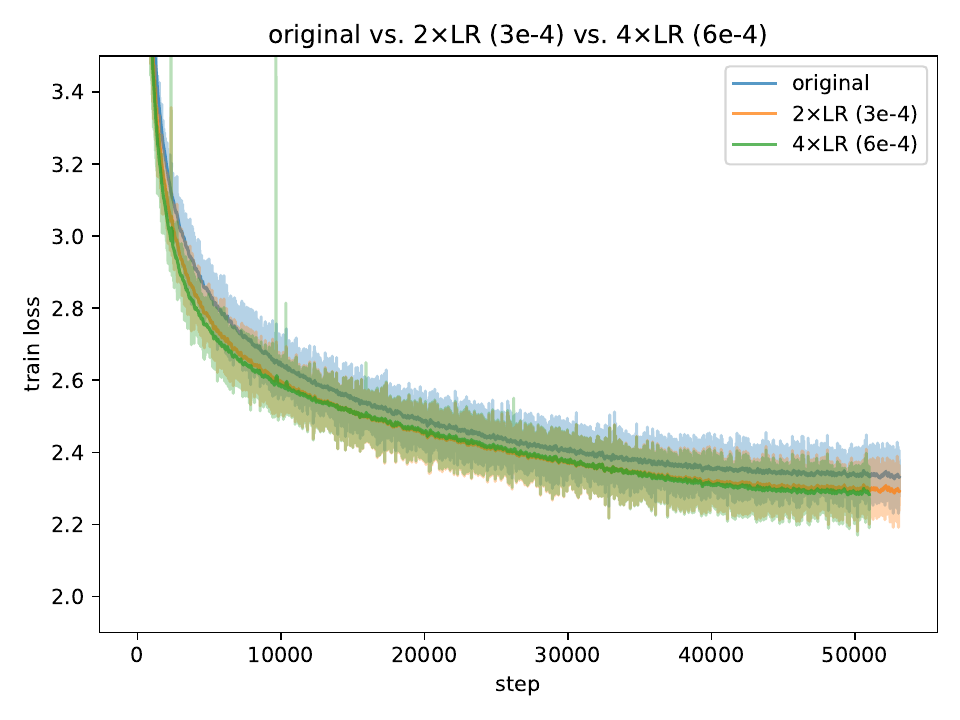}
            \captionof{figure}{``2×/4×~base~LR'' ablation}\label{fig:training-factors-times-base-lr}
        \end{minipage}
    \end{minipage}
        \vspace*{0.3cm} %

    \caption{Training loss curves using (a)~RMSNorm, (b)~no position embedding, (c)~Attention with linear biases position embedding, (d)~grouped-query Attention with 2~groups, (e)~the Adan optimizer, and (f)~2×~or 4×~the base learning rate. All plotted against the baseline.}\label{fig:training-ablations-plots-2}
\end{figure*}

We present here the training ablations and our methodology in more detail. Table~\ref{tab:training-ablations} summarizes the main results, while Figures~\ref{fig:training-ablations-plots} and~\ref{fig:training-ablations-plots-2} display training loss curves of the various ablations against the baseline (``original'' in the plots). The loss curves are throughput-normalized where appropriate.

We conducted the experiments on 2.6B scale, and focused on 3~downstream-tasks:
(1)~ARC~Easy~\cite{DBLP:journals/corr/abs-1803-05457}, (2)~HellaSwag~\cite{DBLP:conf/acl/ZellersHBFC19}, and (3)~LAMBADA~\cite{paperno2016lambada}.
Because the models often achieved better-than-random performance in these evaluation tasks and because they test different abilities of the model, they were deemed a good-enough proxy measure of general downstream improvements.

Consistent throughput across runs is critical in comparing in the compute-equivalent setting. While guaranteeing consistent throughput across runs was not possible due to external factors (such as nondeterministic job layout on the super-computing cluster, or varied I/O throughput due to a shared parallel file system), we believe the variance to be small enough (up to a percent) not to significantly affect our findings. Due to encountering errors as well as resource constraints, we had to cut several ablations short. Because we believe our findings to be meaningful nonetheless, we include unfinished ablations as well.

We assume that every training iteration takes the average amount of time and thus apply the same proportional normalization to every step. This assumption is fine for the conducted ablations because there is no theoretical throughput variance between steps.
We normalize across time with regard to the baseline to compare in the compute-equivalent setting using the following formula:
\begin{equation*} \label{eq:time-normalization}
    t_{i}(s) = s \frac{1}{\bar{T}_{i}},
\end{equation*}
\begin{equation*}
    s'_{i} = \operatorname{time-normalize}_{i}(s) = s \frac{t_{i}(s)}{t_{\text{baseline}}(s)},
\end{equation*}
where $t$~is a time (e.g., in seconds), $i$~is one of the ablation experiments (including the baseline), $s$~is the iteration~(step) to normalize time until, $\bar{T}$~are averaged throughput values (e.g., iterations per second), and $s'_{i}$~is a time-normalized step for experiment~$i$.
This normalization allows us to compare the training and validation loss at each point in time and express that a method with lower loss at the same point in time as the baseline is more compute-efficient up to that point in time with regard to that loss.

The following ablations were conducted:
\begin{enumerate}[nosep] %
    \item \textbf{SwiGLU:} replace the first layer of the MLP part of the Transformer with a T5-style (i.e., without biases)~\cite{raffel2020exploring,google2020t5v1.1} Swish-activated~\cite{ramachandran2017searching} gated linear unit layer~\cite{dauphin2017language,shazeer2020glu}.
    \item \textbf{Untied in/out embedding:} learn separate weights for the input embedding and output ``unembedding'' layers~\cite{google2020t5v1.1}.
    \item \textbf{No \texttt{Linear} biases:} remove all bias terms in \texttt{Linear} layers~\cite{google2020t5v1.1}. The ablation ran until 33\,000~steps and was not evaluated due to being deemed too far from completion.
    \item \textbf{No GPT-like weight init:} whether to scale weight initialization in layers that a residual path leads into as a function of depth~\cite{radford2019language}. The ablation ran until 39\,000~steps and was not evaluated due to being deemed too far from completion.
    \item \textbf{Head scaling:} multiply each Attention head's output by a learned scalar factor~\cite{shleifer2021normformer}.
    \item \textbf{No dropout:} disable dropout~\cite{srivastava2014dropout} in all layers~\cite{google2020t5v1.1}. The latest checkpoint available for evaluation was at 21\,000~steps. While it showed a strongly monotonic improvement over the baseline, it is especially hard to interpret improvement in training loss in the dropout vs.\ no dropout setting. We evaluated this change even though it was far from finishing training because its improvements were so drastic.
    \item \textbf{RMSNorm:} replace LayerNorm normalization layers~\cite{lei2016layer} with root mean square layer normalization layers~\cite{zhang2019root}.
    \item \textbf{NoPE:} no position embedding~\cite{kazemnejad2024impact}; completely remove position embeddings.
    \item \textbf{ALiBi:} replace the Rotary position embedding~\cite{DBLP:journals/ijon/SuALPBL24} with the Attention with linear biases position embedding~\cite{press2021train}. Because we were unable to use an optimized ALiBi kernel for this ablation, throughput-normalization was not considered out of fairness.
    \item \textbf{GQA~(2~groups):} replace multi-head Attention with grouped-query Attention~\cite{DBLP:conf/emnlp/AinslieLJZLS23} with 2~groups (i.e., 2~key/value heads).
    \item \textbf{Adan~(4×~base~LR):} replace the AdamW optimizer with the Adan optimizer~\cite{xie2024adan}, using the baseline's learning rate multiplied by~4. The increased learning rate was chosen based on previous small-scale experiments.
    \item \textbf{2×/4×~base~LR:} use the baseline's learning rate multiplied by~2 or~4. The ablation with a factor~4 increase did not run until completion; its checkpoint used for evaluation was saved at 48,000~steps, 5\,100~steps before the end of training, and was deemed ``close enough'' to completion to provide a fair evaluation comparison, especially due to its noticeable training loss improvements.
\end{enumerate}

\begin{table*}[htbp]
\centering
\begin{tabular}{lrrrl}
\toprule
Change & ARC Easy & HellaSwag & LAMBADA & Interpretation \\
\midrule
- & 0.535 & 0.355 & 0.503 & 0 \\
SwiGLU & 0.527 & \underline{0.361} & \underline{0.507} & + \\
Untied in/out embedding & 0.524 & 0.355 & 0.498 & - \\
No \texttt{Linear} biases (33k~st.) & - & - & - & +? \\
No GPT-like weight init (39k~st.) & - & - & - & - \\
Head scaling & 0.527 & \underline{0.356} & 0.493 & - \\
\emph{No dropout (21k~st.)} & \emph{0.492} & \emph{0.334} & \emph{0.414} & +? \\
RMSNorm & 0.530 & \underline{0.358} & 0.502 & 0 \\
NoPE & 0.516 & 0.351 & 0.486 & - \\
ALiBi & 0.527 & 0.349 & 0.486 & - \\
GQA (2~groups) & 0.513 & 0.346 & 0.459 & ? \\
Adan (4×~base~LR) & \underline{0.544} & \underline{\textbf{0.374}} & \underline{\textbf{0.522}} & +? \\
2×~learning rate & \underline{0.540} & \underline{0.369} & \underline{0.514} & + \\
\emph{4×~learning rate (48k~st.)} & \textbf{\underline{\emph{0.545}}} & \underline{\emph{0.371}} & \underline{\emph{0.517}} & + \\
\bottomrule
\end{tabular}
\caption{Selected evaluation results. Bold is best, underlined is better than baseline. Italic means the run was evaluated before finishing. Ablations without values were not evaluated. The rightmost column contains a subjective interpretation/recommendation, where ``+'', ``-'', ``0'' indicate a positive, negative, and neutral interpretation, respectively.``?'' indicates it is difficult to make a conclusive statement.}\label{tab:training-ablations}
\end{table*}

We decided to implement most of the ``free lunch'' improvements and some neutral results based around current research results at that point in time while disregarding some findings that were deemed too experimental and/or risky.

Notably, we decided to use neither Adan nor 4×~the learning rate despite these changes yielding the best results. Instead, to minimize risks, we opted to use the same learning rate as Llama-2~\cite{DBLP:journals/corr/abs-2307-09288}. Due to our inexperience at training at this scale, we were worried about possible convergence problems if we had chosen these important parameters incorrectly.

In summary, the chosen changes are: SwiGLU, no biases, no dropout during pre-training, RMSNorm, GQA~(with 2~groups). GQA~was chosen despite its comparatively worse results to optimize for inference in the post-training setting.

\begin{table}[ht]
\centering
\begin{tabular}{ll}
\toprule
\textbf{Hyper-Parameter}         & \textbf{Value}    \\ 
\midrule
Training Objective               & CLM               \\ 
Activation Function              & SwiGLU            \\ 
Seq Length                       & 4096              \\ 
Position Embeddings              & Rotary            \\ 
Num Layers                       & 32                \\ 
Hidden Size                      & 4096              \\ 
FFN Hidden Size                  & 13440             \\ 
Num Attention Heads              & 32                \\ 
Head Dim                         & 128               \\ 
Group Query Attention            & yes               \\ 
Num Query Groups                 & 2                 \\ 
Normalization                    & RMSNorm           \\ 
Learning rate                    & 3e-4              \\ 
Min learning rate                & 1.5e-5              \\ 
Disable bias in linear           & yes               \\ 
Hidden dropout                   & 0.0               \\ 
Attention dropout                & 0.0               \\ 
Optimizer                        & AdamW             \\ 
Beta1                            & 0.9               \\ 
Beta2                            & 0.95              \\ 
Data-type                        & bf16              \\ 
\bottomrule
\end{tabular}
\caption{Hyper-Parameter Configuration}
\label{table:model_hyper_parameters}
\end{table}

\subsection{Impact of Educational Content}\label{appendix_educational_content}

\begin{figure}[h]
    \centering
    \includegraphics[width=\linewidth]{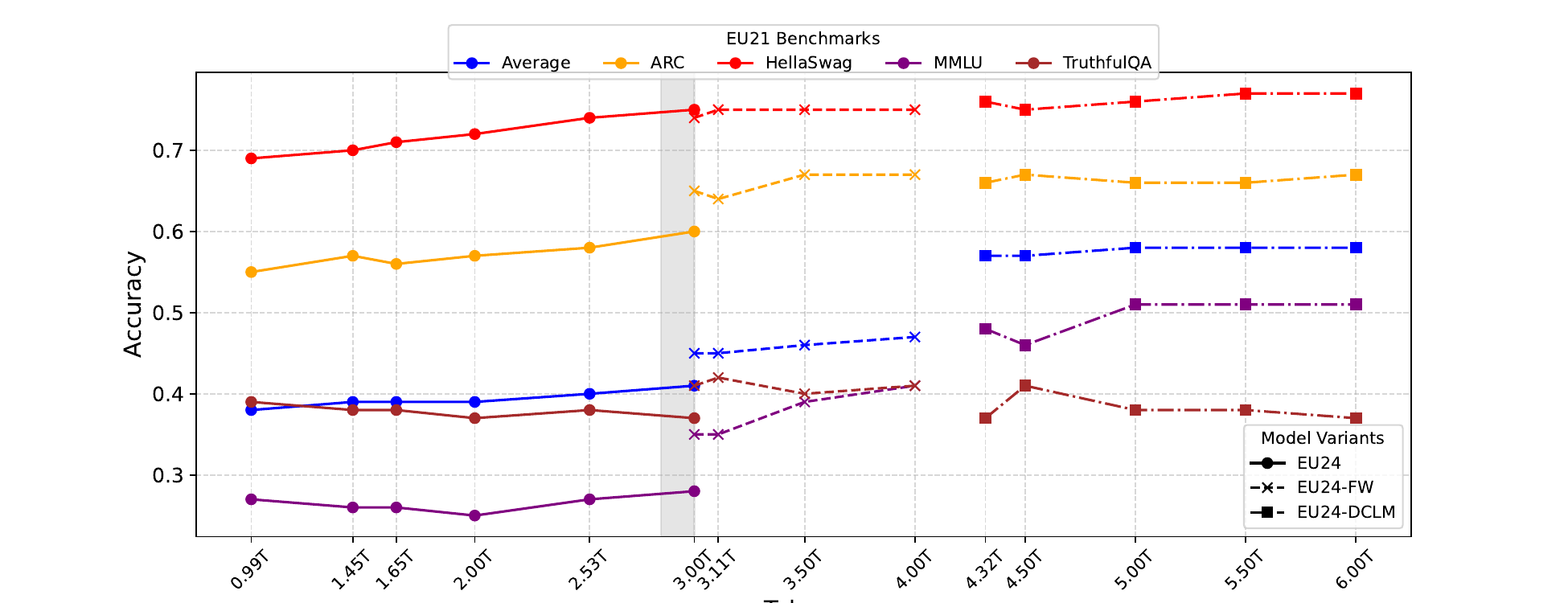}
    \caption{Downstream performance of the base model from 0.99T to 6T tokens for English. The grey area highlights the ablation comparing the performance of EU24 data to the FineWeb-EDU dataset between 2.85T and 3T tokens.  After 4T tokens, we replaced the English FineWeb-EDU and continued training until 6T tokens with DCLM-Baseline.}
    \label{fig:ducational_content_en}
\end{figure}

\begin{figure}[h]
    \centering
    \includegraphics[width=\linewidth]{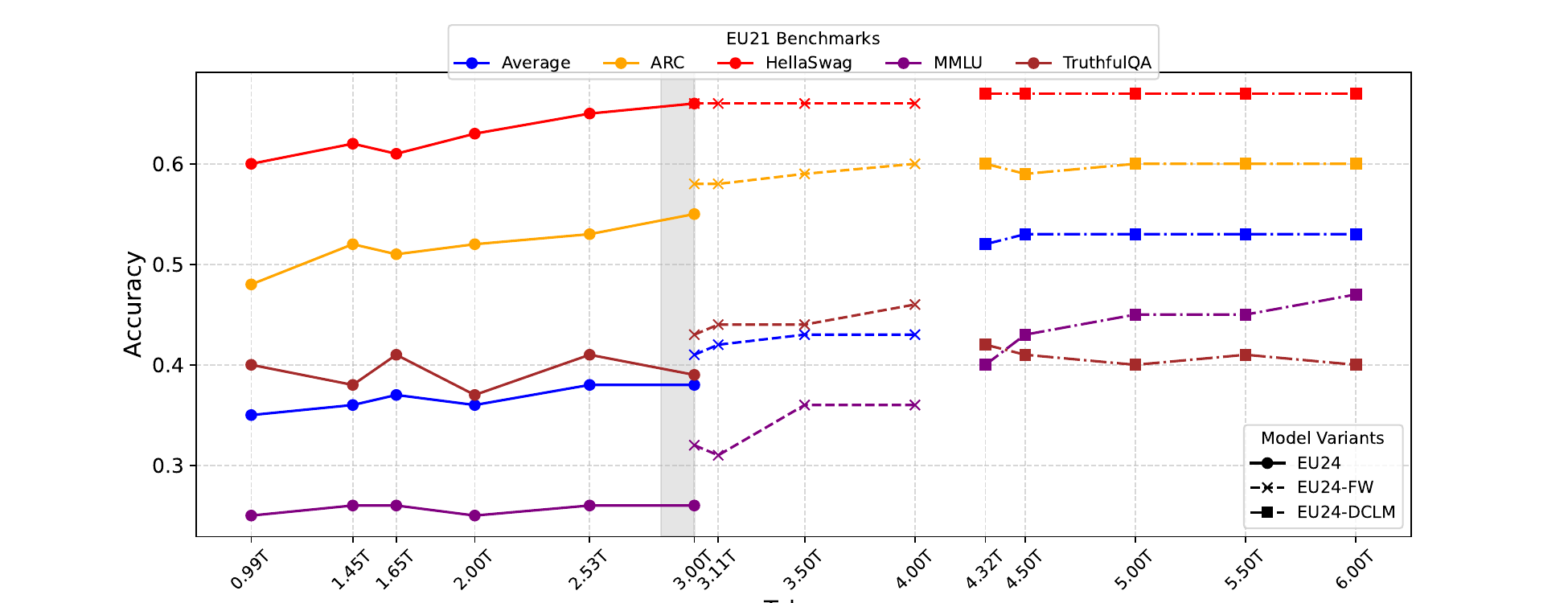}
    \caption{Downstream performance of the base model from 0.99T to 6T tokens for French. The grey area highlights the ablation comparing the performance of EU24 data to the FineWeb-EDU dataset between 2.85T and 3T tokens.  After 4T tokens, we replaced the English FineWeb-EDU and continued training until 6T tokens with DCLM-Baseline.}
    \label{fig:ducational_content_fr}
\end{figure}

\begin{figure}[h]
    \centering
    \includegraphics[width=\linewidth]{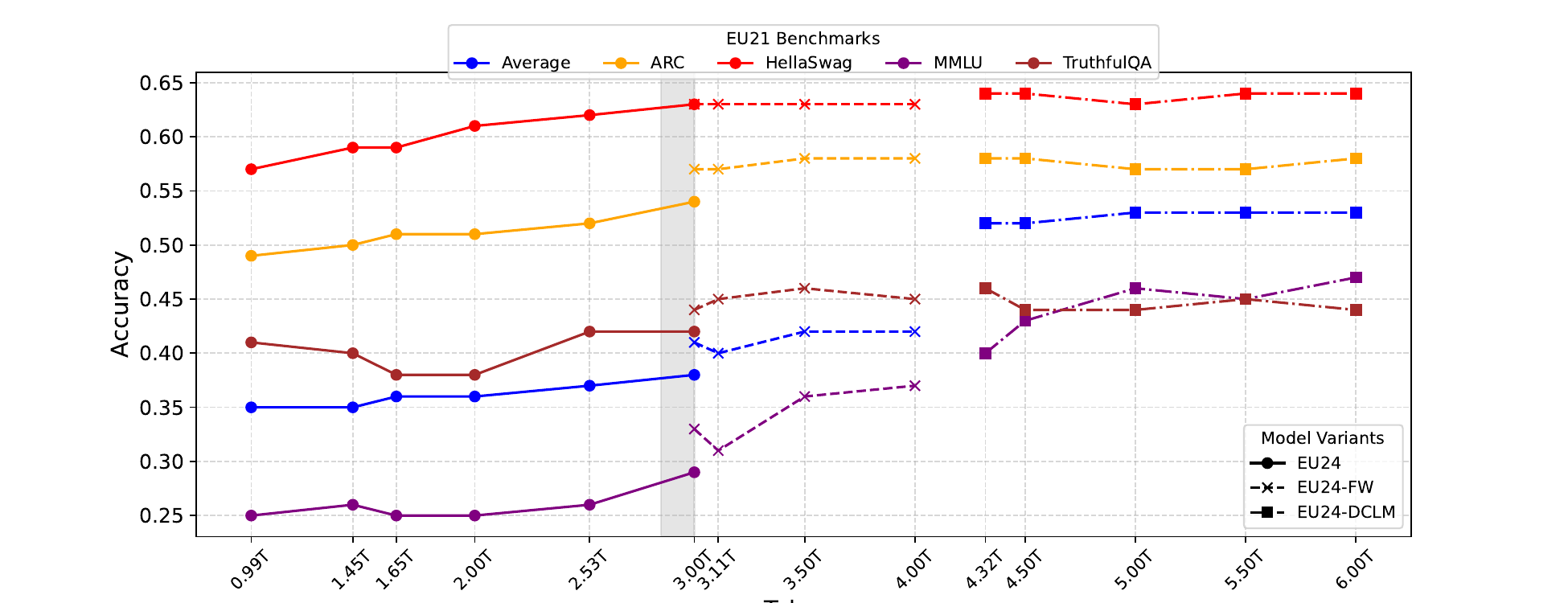}
    \caption{Downstream performance of the base model from 0.99T to 6T tokens for German. The grey area highlights the ablation comparing the performance of EU24 data to the FineWeb-EDU dataset between 2.85T and 3T tokens.  After 4T tokens, we replaced the English FineWeb-EDU and continued training until 6T tokens with DCLM-Baseline.}
    \label{fig:ducational_content_de}
\end{figure}

\begin{figure}[h]
    \centering
    \includegraphics[width=\linewidth]{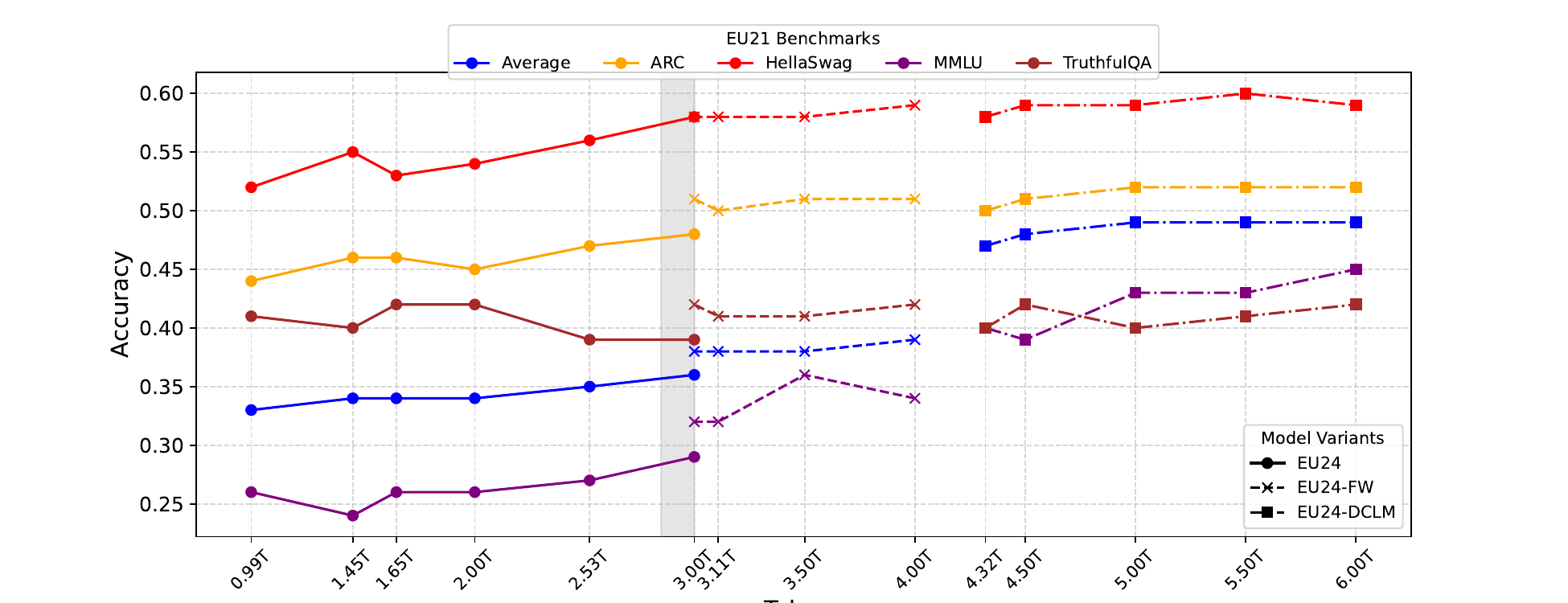}
    \caption{Downstream performance of the base model from 0.99T to 6T tokens for Finish The grey area highlights the ablation comparing the performance of EU24 data to the FineWeb-EDU dataset between 2.85T and 3T tokens.  After 4T tokens, we replaced the English FineWeb-EDU and continued training until 6T tokens with DCLM-Baseline.}
    \label{fig:ducational_content_fi}
\end{figure}

\begin{figure}[h]
    \centering
    \includegraphics[width=\linewidth]{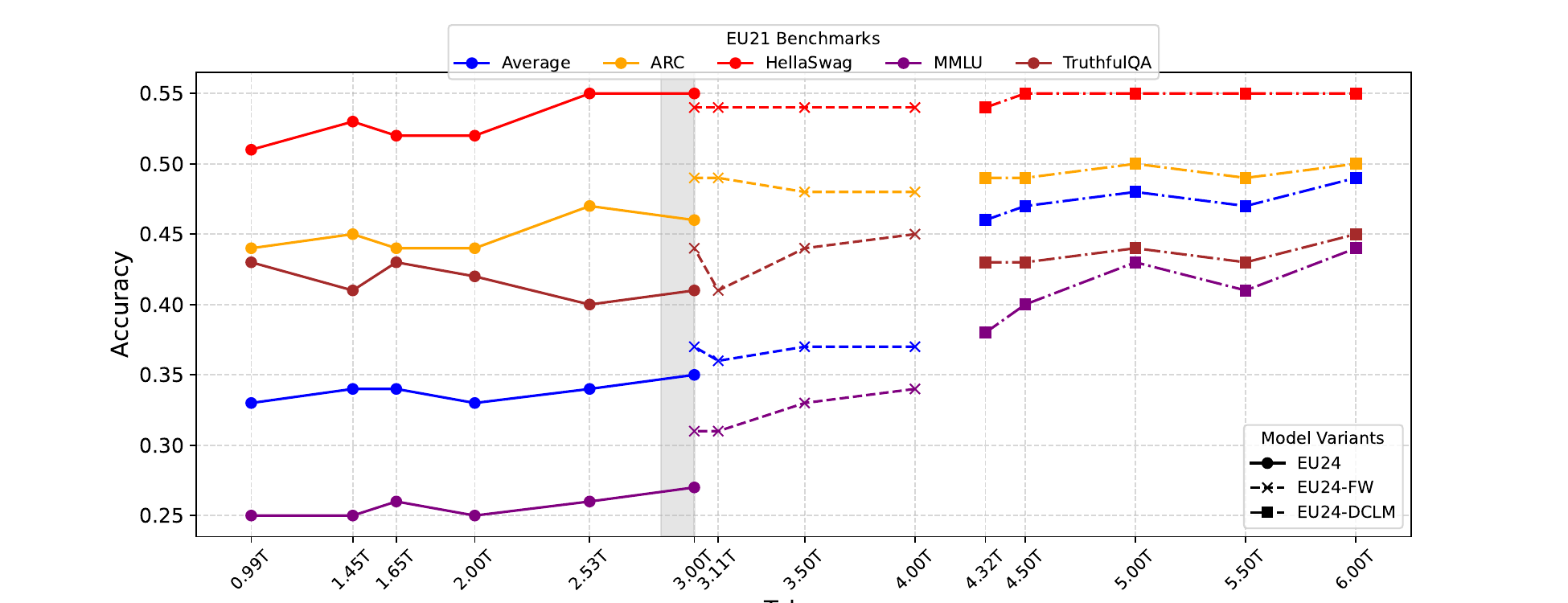}
    \caption{Downstream performance of the base model from 0.99T to 6T tokens for Estonian The grey area highlights the ablation study comparing the performance of EU24 data to the FineWeb-EDU dataset between 2.85T and 3T tokens.  After 4T tokens, we replaced the English FineWeb-EDU and continued training until 6T tokens with DCLM-Baseline.}
    \label{fig:ducational_content_et}
\end{figure}

Figures~\ref{fig:ducational_content_en}-\ref{fig:ducational_content_et} demonstrate the development of the model downstream performance between 0.99T and 6T tokens for English, French, German, Finnish, and Estonian. 
The grey area in the figures highlights the ablation in which we compare the performance of the EU24 data to the FineWeb-EDU dataset between 2.85T and 3T tokens.

\subsection{Evaluation}

The Tables~\ref{tab:21_languages_base}-\ref{tab:11_languages_base} present the results of our base model compared to related base models.

\begin{table*}[t]  
    \centering
    \begin{tabular}{lccccc}
            \toprule
            Model & Average & EU21-ARC & EU21-HeSw & EU21-TQA & EU21-MMLU \\ 
            \hline
            Meta-Llama-3.1-8B & .548 & .554 & .588 & .495 & .556 \\
            Salamandra-7B & .523 & .589 & .637 & .449 & .417 \\
            Mistral-7B-v0.3 & .505 & .513 & .534 & .472 & .501 \\
            Occiglot-7B-eu5 & .464 & .470 & .511 & .448 & .426 \\
            Pharia-1-LLM-7B-control & .409 & .393 & .433 & .456 & .353 \\
            Bloom-7B1 & .348 & .319  & .355 & .464 & .256 \\
            \midrule
            Teuken-7B-Base (Ours) & .520 & .558  & .619  & .449 & .453 \\
            \bottomrule
    \end{tabular}
    \caption{Results on multilingual benchmarks for 21 European languages with base models.}
    \label{tab:21_languages_base}
\end{table*}

\begin{table*}[t]
        \centering
        \begin{tabular}{lccccc}
            \toprule
            Model & Average & EU21-ARC & EU21-HeSw & EU21-TQA & EU21-MMLU \\ 
            \midrule
            Meta-Llama-3.1-8B & .601 & .634  & .687  & .474 & .607  \\
            Mistral-7B-v0.3 & .580 & .632  & .662  & .453 & .571  \\
            Occiglot-7B-eu5 & .568 & .630 & .702  & .424  & .514  \\
            Pharia-1-LLM-7B & .559 & .635  & .686 & .453  & .461  \\
            Salamandra-7B & .545 & .631 & .685  & .429  & .435  \\
            Bloom-7B1 & .406 & .448 & .501  & .416  & .260  \\
            \midrule
            Teuken-7B-Base (Ours) & .541 & .608 & .674  & .405  & .478  \\
            \bottomrule
        \end{tabular}
        \caption{Results on multilingual benchmarks for 6 Languages with base models.}
        \label{tab:6_languages_base}
\end{table*}

 \begin{table*}[t]
        \centering
        \begin{tabular}{lccccc}
            \toprule
            Model & Average & EU21-ARC & EU21-HeSw & EU21-TQA & EU21-MMLU \\ 
            \midrule
            Meta-Llama-3.1-8B & .527 & .522 & .549  & .503 & .535 \\
            Salamandra-7B & .514 & .572 & .617  & .457 & .410 \\
            Mistral-7B-v0.3 & .475 & .466 & .482  & .479 & .473 \\
            Occiglot-7B-eu5 & .422 & .406 & .435  & .458 & .391 \\
            Pharia-1-LLM-7B & .349 & .296 & .332 & .457 & .310   \\
            Bloom-7B1 & .325 & .267 & .296 & .483 & .254 \\
            \midrule
            Teuken-7B-Base (Ours) & .511 & .539 & .597 & .466 & .443 \\
            \bottomrule
        \end{tabular}
        \caption{Results on multilingual benchmarks across 15 exclusive European languages with base models.}
        \label{tab:15_languages_base}
\end{table*}

\begin{table*}[t]
    \centering
    \begin{tabular}{lccccc}
            \toprule
            Model & Average & EU21-ARC & EU21-HeSw & EU21-TQA & EU21-MMLU \\ 
            \midrule
            Salamandra-7B & .536 & .612  & .664  & .447 & .423  \\
            Bloom-7B1 & .376 & .376  & .419  & .452  & .258  \\
            \midrule
            Teuken-7B-Base (Ours) & .533 & .587  & .650  & .432  & .464  \\
            \bottomrule
    \end{tabular}
    \caption{Base model results on multilingual benchmarks across the 10 common languages (Czech, Dutch, English, French, Greek, Italian, Polish, Portuguese, Romanian, and Spanish). The tables include mean accuracy across languages.}
    \label{tab:11_languages_base}
\end{table*}

\begin{table*}[t]
    \centering
    \begin{tabular}{lccccc}
        \toprule
        Model & Average & EU21-ARC & EU21-HeSw & EU21-MMLU & EU21-TQA  \ \\
        \midrule
        Meta-Llama-3.1-8B-Instruct & .591 & .610 & .621 & .615 & .520 \\
        Mistral-7B-Instruct-v0.3 & .584 & .620 & .609 & .540 & .568 \\
        Meta-Llama-3.1-8B & .577 & .589 & .632 & .590 & .495 \\
        Occiglot-7B-eu5-Instruct & .575 & .625 & .691 & .514 & .469 \\
        Pharia-1-LLM-7B-ctr-aligned & .567 & .618 & .669 & .482 & .498 \\
        Aya-23-8B & .564 & .589 & .645 & .519 & .501 \\
        Occiglot-7B-eu5 & .561 & .615 & .679 & .512 & .439 \\
        Mistral-7B-v0.3 & .552 & .595 & .598 & .551 & .465 \\
        Pharia-1-LLM-7B-ctr & .546 & .616 & .659 & .450 & .457 \\
        Salamandra-7B-Instruct & .543 & .614 & .644 & .466 & .451 \\
        Salamandra-7B & .531 & .601 & .645 & .432 & .445 \\
        Bloomz-7B1 & .340 & .299 & .307 & .309 & .442 \\
        Bloom-7B1 & .323 & .294 & .310 & .262 & .427 \\
        \midrule
        Teuken-7B-Instruct (Ours) & .588 & .607 & .689 & .477 & .577 \\
        Teuken-7B-Base (Ours) & .530 & .576 & .636 & .470 & .436 \\
        \bottomrule
    \end{tabular}
    \caption{Task accuracies for the German language.}
    \label{tab:de}
\end{table*}

\begin{table*}[t]
    \centering
    \begin{tabular}{lccccc}
        \toprule
        Model & Average & EU21-ARC & EU21-HeSw & EU21-MMLU & EU21-TQA \ \\
        \midrule
        Mistral-7B-Instruct-v0.3 & .706 & .753 & .846 & .627 & .597 \\
        Meta-Llama-3.1-8B-Instruct & .692 & .736 & .802 & .690 & .541 \\
        Meta-Llama-3.1-8B & .662 & .715 & .819 & .661 & .452 \\
        Mistral-7B-v0.3 & .654 & .727 & .829 & .635 & .426 \\
        Occiglot-7B-eu5-Instruct & .624 & .698 & .797 & .551 & .449 \\
        Occiglot-7B-eu5 & .605 & .681 & .790 & .543 & .404 \\
        Aya-23-8B & .617 & .670 & .779 & .565 & .454 \\
        Pharia-1-LLM-7B-ctr-aligned & .623 & .712 & .775 & .519 & .486 \\
        Pharia-1-LLM-7B-ctr & .607 & .702 & .770 & .503 & .454 \\
        Salamandra-7B-Instruct & .609 & .711 & .772 & .508 & .446 \\
        Salamandra-7B & .586 & .695 & .771 & .453 & .423 \\
        Bloomz-7B1 & .496 & .550 & .610 & .372 & .452 \\
        Bloom-7B1 & .449 & .546 & .606 & .257 & .389 \\
        \midrule
        Teuken-7B-Instruct (Ours) & .635 & .685 & .820 & .516 & .517 \\
        Teuken-7B-Base (Ours) & .581 & .667 & .771 & .512 & .372 \\
        \bottomrule
    \end{tabular}
    \caption{Task accuracies for the English language.}
    \label{tab:en}
\end{table*}

\begin{table*}[t]
    \centering
    \begin{tabular}{lccccc}
        \toprule
        Model & Average & EU21-ARC & EU21-HeSw & EU21-MMLU & EU21-TQA \ \\
        \midrule
        Meta-Llama-3.1-8B-Instruct & .612 & .638 & .667 & .624 & .521 \\
        Mistral-7B-Instruct-v0.3 & .606 & .647 & .660 & .549 & .567 \\
        Meta-Llama-3.1-8B & .593 & .620 & .675 & .598 & .478 \\
        Occiglot-7B-eu5-Instruct & .581 & .656 & .722 & .511 & .435 \\
        Pharia-1-LLM-7B-ctr-aligned & .575 & .643 & .698 & .477 & .482 \\
        Mistral-7B-v0.3 & .573 & .627 & .652 & .562 & .451 \\
        Aya-23-8B & .564 & .610 & .687 & .519 & .441 \\
        Occiglot-7B-eu5 & .563 & .643 & .713 & .504 & .393 \\
        Salamandra-7B-Instruct & .555 & .635 & .681 & .461 & .444 \\
        Pharia-1-LLM-7B-ctr & .551 & .636 & .684 & .458 & .428 \\
        Salamandra-7B & .532 & .624 & .683 & .430 & .391 \\
        Bloomz-7B1 & .461 & .500 & .579 & .367 & .396 \\
        Bloom-7B1 & .431 & .501 & .573 & .262 & .387 \\
        \midrule
        Teuken-7B-Instruct (Ours) & .592 & .634 & .710 & .476 & .550 \\
        Teuken-7B-Base (Ours) & .535 & .602 & .673 & .466 & .399 \\
        \bottomrule
    \end{tabular}
    \caption{Task accuracies for the French language.}
    \label{tab:fr}
\end{table*}

\begin{table*}[t]
    \centering
    \begin{tabular}{lccccc}
        \toprule
        Model & Average & EU21-ARC & EU21-HeSw & EU21-MMLU & EU21-TQA \ \\
        \midrule
        Meta-Llama-3.1-8B-Instruct & .616 & .635 & .649 & .614 & .565 \\
        Mistral-7B-Instruct-v0.3 & .590 & .625 & .627 & .545 & .565 \\
        Occiglot-7B-eu5-Instruct & .589 & .647 & .708 & .519 & .483 \\
        Meta-Llama-3.1-8B & .588 & .624 & .655 & .592 & .483 \\
        Pharia-1-LLM-7B-ctr-aligned & .575 & .627 & .679 & .480 & .515 \\
        Occiglot-7B-eu5 & .575 & .629 & .702 & .517 & .451 \\
        Aya-23-8B & .565 & .596 & .663 & .513 & .488 \\
        Mistral-7B-v0.3 & .561 & .601 & .618 & .556 & .471 \\
        Salamandra-7B-Instruct & .558 & .626 & .658 & .472 & .478 \\
        Pharia-1-LLM-7B-ctr & .554 & .616 & .670 & .457 & .473 \\
        Salamandra-7B & .536 & .610 & .656 & .432 & .448 \\
        Bloomz-7B1 & .368 & .339 & .380 & .326 & .428 \\
        Bloom-7B1 & .363 & .345 & .390 & .263 & .453 \\
        \midrule
        Teuken-7B-Instruct (Ours) & .588 & .634 & .688 & .474 & .556 \\
        Teuken-7B-Base (Ours) & .529 & .590 & .646 & .475 & .404 \\
        \bottomrule
    \end{tabular}
    \caption{Task accuracies for the Italian language.}
    \label{tab:it}
\end{table*}

\begin{table*}[t]
    \centering
    \begin{tabular}{lccccc}
        \toprule
        Model & Average & EU21-ARC & EU21-HeSw & EU21-MMLU & EU21-TQA \ \\
        \midrule
        Meta-Llama-3.1-8B-Instruct & .618 & .641 & .673 & .627 & .530 \\
        Mistral-7B-Instruct-v0.3 & .602 & .655 & .654 & .548 & .552 \\
        Meta-Llama-3.1-8B & .597 & .632 & .679 & .606 & .470 \\
        Occiglot-7B-eu5-Instruct & .591 & .656 & .729 & .520 & .458 \\
        Pharia-1-LLM-7B-ctr-aligned & .578 & .648 & .693 & .480 & .492 \\
        Occiglot-7B-eu5 & .576 & .636 & .712 & .516 & .438 \\
        Mistral-7B-v0.3 & .574 & .628 & .651 & .561 & .456 \\
        Salamandra-7B-Instruct & .566 & .643 & .687 & .483 & .450 \\
        Aya-23-8B & .566 & .608 & .677 & .523 & .456 \\
        Pharia-1-LLM-7B-ctr & .552 & .636 & .681 & .453 & .437 \\
        Salamandra-7B & .549 & .634 & .686 & .434 & .443 \\
        Bloomz-7B1 & .478 & .517 & .582 & .379 & .434 \\
        Bloom-7B1 & .437 & .509 & .573 & .260 & .407 \\
        \midrule
        Teuken-7B-Instruct (Ours) & .597 & .647 & .710 & .474 & .557 \\
        Teuken-7B-Base (Ours) & .541 & .620 & .666 & .473 & .404 \\
        \bottomrule
    \end{tabular}
    \caption{Task accuracies for the Spanish language.}
    \label{tab:es}
\end{table*}

\begin{table*}[t]
    \centering
    \begin{tabular}{lccccc}
        \toprule
        Model & Average & EU21-ARC & EU21-HeSw & EU21-MMLU & EU21-TQA \ \\
        \midrule
        Meta-Llama-3.1-8B-Instruct & .608 & .625 & .651 & .622 & .535 \\
        Mistral-7B-Instruct-v0.3 & .590 & .625 & .625 & .551 & .558 \\
        Meta-Llama-3.1-8B & .587 & .624 & .662 & .596 & .466 \\
        Aya-23-8B & .570 & .608 & .674 & .517 & .479 \\
        Pharia-1-LLM-7B-ctr-aligned & .563 & .611 & .661 & .471 & .508 \\
        Mistral-7B-v0.3 & .562 & .614 & .624 & .561 & .449 \\
        Salamandra-7B-Instruct & .557 & .631 & .667 & .473 & .459 \\
        Pharia-1-LLM-7B-ctr & .542 & .603 & .649 & .448 & .468 \\
        Occiglot-7B-eu5-Instruct & .541 & .593 & .623 & .490 & .457 \\
        Salamandra-7B & .535 & .619 & .669 & .430 & .422 \\
        Occiglot-7B-eu5 & .526 & .578 & .615 & .491 & .420 \\
        Bloomz-7B1 & .455 & .490 & .555 & .373 & .403 \\
        Bloom-7B1 & .433 & .490 & .553 & .255 & .434 \\
        \midrule
        Teuken-7B-Instruct (Ours) & .587 & .636 & .699 & .476 & .538 \\
        Teuken-7B-Base (Ours) & .533 & .593 & .655 & .473 & .412 \\
        \bottomrule
    \end{tabular}
    \caption{Task accuracies for the Portuguese language.}
    \label{tab:pt-pt}
\end{table*}

\begin{table*}[t]
    \centering
    \begin{tabular}{lccccc}
        \toprule
        Model & Average & EU21-ARC & EU21-HeSw & EU21-MMLU & EU21-TQA \ \\
        \midrule
        Meta-Llama-3.1-8B-Instruct & .583 & .589 & .589 & .596 & .558 \\
        Meta-Llama-3.1-8B & .568 & .577 & .597 & .580 & .518 \\
        Mistral-7B-Instruct-v0.3 & .563 & .584 & .550 & .519 & .597 \\
        Aya-23-8B & .559 & .578 & .638 & .505 & .516 \\
        Mistral-7B-v0.3 & .546 & .558 & .550 & .537 & .540 \\
        Salamandra-7B-Instruct & .546 & .598 & .629 & .465 & .492 \\
        Salamandra-7B & .533 & .595 & .634 & .418 & .485 \\
        Occiglot-7B-eu5-Instruct & .490 & .516 & .499 & .433 & .510 \\
        Occiglot-7B-eu5 & .474 & .494 & .488 & .438 & .477 \\
        Pharia-1-LLM-7B-ctr-aligned & .377 & .322 & .336 & .352 & .497 \\
        Pharia-1-LLM-7B-ctr & .372 & .314 & .333 & .350 & .490 \\
        Bloomz-7B1 & .346 & .287 & .302 & .299 & .497 \\
        Bloom-7B1 & .341 & .287 & .302 & .258 & .518 \\
        \midrule
        Teuken-7B-Instruct (Ours) & .578 & .592 & .650 & .470 & .601 \\
        Teuken-7B-Base (Ours) & .525 & .558 & .607 & .457 & .479 \\
        \bottomrule
    \end{tabular}
    \caption{Task accuracies for the Romanian language.}
    \label{tab:ro}
\end{table*}

\begin{table*}[t]
    \centering
    \begin{tabular}{lccccc}
        \toprule
        Model & Average & EU21-ARC & EU21-HeSw & EU21-MMLU & EU21-TQA \ \\
        \midrule
        Meta-Llama-3.1-8B-Instruct & .565 & .577 & .571 & .581 & .529 \\
        Meta-Llama-3.1-8B & .553 & .566 & .583 & .556 & .508 \\
        Mistral-7B-Instruct-v0.3 & .550 & .580 & .545 & .510 & .563 \\
        Aya-23-8B & .536 & .569 & .611 & .499 & .467 \\
        Salamandra-7B-Instruct & .536 & .591 & .624 & .461 & .467 \\
        Mistral-7B-v0.3 & .528 & .564 & .540 & .520 & .487 \\
        Salamandra-7B & .512 & .587 & .627 & .405 & .429 \\
        Occiglot-7B-eu5-Instruct & .466 & .484 & .485 & .435 & .463 \\
        Occiglot-7B-eu5 & .456 & .461 & .477 & .436 & .449 \\
        Bloomz-7B1 & .334 & .267 & .296 & .287 & .488 \\
        Pharia-1-LLM-7B-ctr & .332 & .275 & .307 & .302 & .444 \\
        Pharia-1-LLM-7B-ctr-aligned & .330 & .270 & .310 & .306 & .433 \\
        Bloom-7B1 & .327 & .280 & .298 & .259 & .472 \\
        \midrule
        Teuken-7B-Instruct (Ours) & .567 & .602 & .649 & .458 & .557 \\
        Teuken-7B-Base (Ours) & .518 & .562 & .609 & .450 & .453 \\
        \bottomrule
    \end{tabular}
    \caption{Task accuracies for the Czech language.}
    \label{tab:cs}
\end{table*}

\begin{table*}[t]
    \centering
    \begin{tabular}{lccccc}
        \toprule
        Model & Average & EU21-ARC & EU21-HeSw & EU21-MMLU & EU21-TQA \ \\
        \midrule
        Meta-Llama-3.1-8B-Instruct & .576 & .571 & .605 & .596 & .533 \\
        Mistral-7B-Instruct-v0.3 & .564 & .582 & .589 & .532 & .555 \\
        Meta-Llama-3.1-8B & .562 & .556 & .616 & .569 & .506 \\
        Salamandra-7B-Instruct & .541 & .601 & .663 & .462 & .436 \\
        Mistral-7B-v0.3 & .535 & .561 & .580 & .542 & .457 \\
        Salamandra-7B & .522 & .594 & .659 & .425 & .410 \\
        Occiglot-7B-eu5-Instruct & .490 & .500 & .541 & .452 & .466 \\
        Occiglot-7B-eu5 & .472 & .484 & .529 & .448 & .429 \\
        Aya-23-8B & .441 & .404 & .476 & .440 & .444 \\
        Pharia-1-LLM-7B-ctr-aligned & .364 & .319 & .358 & .364 & .416 \\
        Pharia-1-LLM-7B-ctr & .358 & .325 & .354 & .339 & .412 \\
        Bloomz-7B1 & .330 & .268 & .306 & .306 & .440 \\
        Bloom-7B1 & .322 & .270 & .306 & .253 & .461 \\
        \midrule
        Teuken-7B-Instruct (Ours) & .578 & .597 & .692 & .471 & .555 \\
        Teuken-7B-Base (Ours) & .528 & .566 & .642 & .467 & .437 \\
        \bottomrule
    \end{tabular}
    \caption{Task accuracies for the Danish language.}
    \label{tab:da}
\end{table*}

\begin{table*}[t]
    \centering
    \begin{tabular}{lccccc}
        \toprule
        Model & Average & EU21-ARC & EU21-HeSw & EU21-MMLU & EU21-TQA \ \\
        \midrule
            Aya-23-8B & .541 & .556 & .622 & .478 & .506 \\
            Salamandra-7B-Instruct & .537 & .574 & .633 & .425 & .514 \\
            Meta-Llama-3.1-8B-Instruct & .530 & .510 & .540 & .536 & .535 \\
            Meta-Llama-3.1-8B & .519 & .504 & .553 & .513 & .504 \\
            Salamandra-7B & .518 & .570 & .633 & .387 & .480 \\
            Mistral-7B-Instruct-v0.3 & .379 & .309 & .376 & .362 & .470 \\
            Mistral-7B-v0.3 & .377 & .321 & .383 & .364 & .440 \\
            Occiglot-7B-eu5-Instruct & .350 & .302 & .354 & .311 & .434 \\
            Occiglot-7B-eu5 & .350 & .289 & .354 & .315 & .440 \\
            Pharia-1-LLM-7B-ctr-aligned & .325 & .251 & .289 & .275 & .483 \\
            Pharia-1-LLM-7B-ctr & .323 & .247 & .290 & .275 & .481 \\
            Bloom-7B1 & .323 & .254 & .287 & .254 & .496 \\
            Bloomz-7B1 & .322 & .243 & .285 & .273 & .488 \\
            \midrule
            Teuken-7B-Instruct (Ours) & .541 & .566 & .653 & .367 & .579 \\
            Teuken-7B-Base (Ours) & .515 & .548 & .620 & .429 & .463 \\
        \bottomrule
    \end{tabular}
    \caption{Task accuracies for the Greek language.}
    \label{tab:el}
\end{table*}

\begin{table*}[t]
    \centering
    \begin{tabular}{lccccc}
            \toprule
            Model & Average & EU21-ARC & EU21-HeSw & EU21-MMLU & EU21-TQA  \\
            \midrule
            Salamandra-7B-Instruct & .510 & .541 & .576 & .441 & .480 \\
            Salamandra-7B & .482 & .530 & .574 & .407 & .419 \\
            Meta-Llama-3.1-8B-Instruct & .472 & .443 & .452 & .501 & .492 \\
            Meta-Llama-3.1-8B & .464 & .444 & .463 & .487 & .464 \\
            Mistral-7B-Instruct-v0.3 & .369 & .311 & .338 & .372 & .456 \\
            Mistral-7B-v0.3 & .364 & .293 & .342 & .389 & .431 \\
            Occiglot-7B-eu5-Instruct & .352 & .290 & .334 & .322 & .461 \\
            Occiglot-7B-eu5 & .347 & .287 & .333 & .327 & .442 \\
            Aya-23-8B & .345 & .286 & .327 & .334 & .434 \\
            Pharia-1-LLM-7B-ctr-aligned & .318 & .257 & .295 & .288 & .434 \\
            Pharia-1-LLM-7B-ctr & .316 & .258 & .297 & .286 & .423 \\
            Bloom-7B1 & .315 & .256 & .288 & .248 & .466 \\
            Bloomz-7B1 & .311 & .263 & .285 & .271 & .425 \\
            \midrule
            Teuken-7B-Instruct (Ours) & .535 & .536 & .595 & .440 & .570 \\
            Teuken-7B-Base (Ours) & .485 & .501 & .552 & .441 & .448 \\
            \bottomrule
    \end{tabular}
    \caption{Task accuracies for the Estonian language.}
    \label{tab:Estonian}
\end{table*}

\begin{table*}[t]
    \centering
    \begin{tabular}{lccccc}
            \toprule
            Model & Average & EU21-ARC & EU21-HeSw & EU21-MMLU & EU21-TQA \ \\
            \midrule
            Meta-Llama-3.1-8B-Instruct & .505 & .491 & .511 & .527 & .491 \\
            Meta-Llama-3.1-8B & .493 & .483 & .523 & .512 & .454 \\
            Salamandra-7B-Instruct & .512 & .532 & .602 & .439 & .476 \\
            Salamandra-7B & .491 & .540 & .603 & .415 & .404 \\
            Mistral-7B-Instruct-v0.3 & .410 & .359 & .404 & .416 & .463 \\
            Mistral-7B-v0.3 & .403 & .358 & .406 & .426 & .424 \\
            Occiglot-7B-eu5-Instruct & .374 & .328 & .371 & .351 & .446 \\
            Occiglot-7B-eu5 & .367 & .317 & .368 & .354 & .431 \\
            Aya-23-8B & .356 & .299 & .348 & .349 & .428 \\
            Pharia-1-LLM-7B-ctr-aligned & .325 & .275 & .305 & .304 & .417 \\
            Pharia-1-LLM-7B-ctr & .324 & .281 & .300 & .289 & .427 \\
            Bloomz-7B1 & .307 & .267 & .296 & .269 & .395 \\
            Bloom-7B1 & .313 & .258 & .301 & .247 & .446 \\
            \midrule
            Teuken-7B-Instruct (Ours) & .550 & .559 & .631 & .447 & .564 \\
            Teuken-7B-Base (Ours) & .493 & .520 & .590 & .446 & .415 \\
            \bottomrule
    \end{tabular}
    \caption{Task accuracies for the Finnish language.}
    \label{tab:Finnish}
\end{table*}

\begin{table*}[t]
    \centering
    \begin{tabular}{lccccc}
            \toprule
            Model & Average & EU21-ARC & EU21-HeSw & EU21-MMLU & EU21-TQA \\
            \midrule
            Meta-Llama-3.1-8B-Instruct & .544 & .524 & .558 & .566 & .530 \\
            Meta-Llama-3.1-8B & .533 & .522 & .558 & .547 & .506 \\
            Salamandra-7B-Instruct & .526 & .553 & .592 & .444 & .514 \\
            Mistral-7B-Instruct-v0.3 & .521 & .515 & .504 & .488 & .576 \\
            Salamandra-7B & .502 & .541 & .589 & .400 & .476 \\
            Mistral-7B-v0.3 & .492 & .492 & .496 & .501 & .479 \\
            Occiglot-7B-eu5-Instruct & .434 & .424 & .422 & .398 & .491 \\
            Occiglot-7B-eu5 & .426 & .412 & .417 & .403 & .470 \\
            Aya-23-8B & .378 & .299 & .347 & .374 & .493 \\
            Pharia-1-LLM-7B-ctr-aligned & .334 & .265 & .297 & .297 & .478 \\
            Pharia-1-LLM-7B-ctr & .333 & .265 & .294 & .292 & .478 \\
            Bloomz-7B1 & .336 & .268 & .294 & .272 & .509 \\
            Bloom-7B1 & .332 & .276 & .290 & .244 & .517 \\
            \midrule
            Teuken-7B-Instruct (Ours) & .548 & .546 & .614 & .447 & .585 \\
            Teuken-7B-Base (Ours) & .501 & .526 & .575 & .434 & .471 \\
            \bottomrule
    \end{tabular}
    \caption{Task accuracies for the Hungarian language.}
    \label{tab:Hungarian}
\end{table*}

\begin{table*}[t]
    \centering
    \begin{tabular}{lccccc}
            \toprule
            Model & Average & EU21-ARC & EU21-HeSw & EU21-MMLU & EU21-TQA \\
            \midrule
            Salamandra-7B-Instruct & .527 & .554 & .589 & .444 & .519 \\
            Salamandra-7B & .503 & .557 & .580 & .394 & .481 \\
            Meta-Llama-3.1-8B-Instruct & .470 & .447 & .442 & .498 & .495 \\
            Meta-Llama-3.1-8B & .465 & .444 & .447 & .482 & .486 \\
            Mistral-7B-Instruct-v0.3 & .370 & .292 & .337 & .365 & .485 \\
            Mistral-7B-v0.3 & .366 & .289 & .339 & .374 & .463 \\
            Occiglot-7B-eu5-Instruct & .352 & .283 & .328 & .326 & .472 \\
            Occiglot-7B-eu5 & .349 & .277 & .326 & .327 & .466 \\
            Aya-23-8B & .396 & .318 & .378 & .390 & .499 \\
            Pharia-1-LLM-7B-ctr-aligned & .322 & .248 & .298 & .286 & .456 \\
            Pharia-1-LLM-7B-ctr & .319 & .246 & .300 & .283 & .447 \\
            Bloomz-7B1 & .318 & .254 & .291 & .267 & .459 \\
            Bloom-7B1 & .322 & .261 & .293 & .259 & .476 \\
            \midrule
            Teuken-7B-Instruct (Ours) & .543 & .538 & .605 & .439 & .591 \\
            Teuken-7B-Base (Ours) & .494 & .508 & .556 & .430 & .482 \\
            \bottomrule
    \end{tabular}
    \caption{Task accuracies for the Lithuanian language.}
    \label{tab:Lithuanian}
\end{table*}

\begin{table*}[t]
    \centering
    \begin{tabular}{lccccc}
            \toprule
            Model & Average & EU21-ARC & EU21-HeSw & EU21-MMLU & EU21-TQA \\
            \midrule
            Meta-Llama-3.1-8B-Instruct & .469 & .432 & .438 & .488 & .518 \\
            Meta-Llama-3.1-8B & .468 & .432 & .443 & .482 & .515 \\
            Salamandra-7B-Instruct & .514 & .533 & .565 & .433 & .526 \\
            Salamandra-7B & .502 & .525 & .570 & .406 & .506 \\
            Mistral-7B-Instruct-v0.3 & .375 & .295 & .326 & .352 & .525 \\
            Mistral-7B-v0.3 & .373 & .292 & .329 & .368 & .502 \\
            Occiglot-7B-eu5-Instruct & .366 & .294 & .320 & .326 & .522 \\
            Occiglot-7B-eu5 & .358 & .281 & .321 & .324 & .507 \\
            Aya-23-8B & .358 & .283 & .324 & .338 & .486 \\
            Pharia-1-LLM-7B-ctr-aligned & .334 & .267 & .294 & .294 & .480 \\
            Pharia-1-LLM-7B-ctr & .328 & .260 & .296 & .280 & .473 \\
            Bloomz-7B1 & .330 & .261 & .291 & .276 & .494 \\
            Bloom-7B1 & .327 & .259 & .293 & .255 & .503 \\
            \midrule
            Teuken-7B-Instruct (Ours) & .537 & .520 & .585 & .425 & .619 \\
            Teuken-7B-Base (Ours) & .495 & .492 & .549 & .427 & .514 \\
            \bottomrule
    \end{tabular}
    \caption{Task accuracies for the Latvian language.}
    \label{tab:Latvian}
\end{table*}

\begin{table*}[t]
    \centering
    \begin{tabular}{lccccc}
            \toprule
            Model & Average & EU21-ARC & EU21-HeSw & EU21-MMLU & EU21-TQA \\
            \midrule
            Meta-Llama-3.1-8B-Instruct & .597 & .592 & .629 & .608 & .561 \\
            Mistral-7B-Instruct-v0.3 & .582 & .595 & .593 & .534 & .608 \\
            Meta-Llama-3.1-8B & .580 & .583 & .632 & .586 & .520 \\
            Salamandra-7B-Instruct & .560 & .603 & .655 & .471 & .511 \\
            Aya-23-8B & .560 & .576 & .640 & .509 & .513 \\
            Pharia-1-LLM-7B-ctr-aligned & .557 & .595 & .652 & .470 & .510 \\
            Mistral-7B-v0.3 & .553 & .568 & .584 & .546 & .512 \\
            Pharia-1-LLM-7B-ctr & .548 & .595 & .642 & .448 & .505 \\
            Salamandra-7B & .539 & .593 & .648 & .431 & .483 \\
            Occiglot-7B-eu5-Instruct & .512 & .530 & .563 & .468 & .485 \\
            Occiglot-7B-eu5 & .498 & .515 & .549 & .466 & .465 \\
            Bloomz-7B1 & .342 & .264 & .305 & .303 & .494 \\
            Bloom-7B1 & .331 & .277 & .310 & .253 & .486 \\
            \midrule
            Teuken-7B-Instruct (Ours) & .584 & .599 & .685 & .472 & .582 \\
            Teuken-7B-Base (Ours) & .538 & .577 & .644 & .462 & .468 \\
            \bottomrule
    \end{tabular}
    \caption{Task accuracies for the Dutch language.}
    \label{tab:Dutch}
\end{table*}

\begin{table*}[t]
    \centering
    \begin{tabular}{lccccc}
            \toprule
            Model & Average & EU21-ARC & EU21-HeSw & EU21-MMLU & EU21-TQA \\
            \midrule
            Mistral-7B-Instruct-v0.3 & .563 & .591 & .572 & .496 & .593 \\
            Meta-Llama-3.1-8B-Instruct & .547 & .554 & .544 & .546 & .546 \\
            Salamandra-7B-Instruct & .545 & .595 & .642 & .449 & .493 \\
            Mistral-7B-v0.3 & .538 & .563 & .568 & .488 & .531 \\
            Meta-Llama-3.1-8B & .534 & .540 & .559 & .516 & .520 \\
            Salamandra-7B & .519 & .586 & .638 & .389 & .463 \\
            Occiglot-7B-eu5-Instruct & .464 & .470 & .480 & .369 & .538 \\
            Occiglot-7B-eu5 & .448 & .464 & .474 & .344 & .511 \\
            Aya-23-8B & .422 & .386 & .421 & .403 & .477 \\
            Pharia-1-LLM-7B-ctr & .335 & .264 & .305 & .270 & .500 \\
            Pharia-1-LLM-7B-ctr-aligned & .333 & .253 & .310 & .263 & .507 \\
            Bloom-7B1 & .332 & .259 & .295 & .259 & .517 \\
            Bloomz-7B1 & .326 & .260 & .292 & .251 & .500 \\
            \midrule
            Teuken-7B-Instruct (Ours) & .557 & .581 & .661 & .387 & .597 \\
            Teuken-7B-Base (Ours) & .515 & .545 & .614 & .410 & .492 \\
            \bottomrule
    \end{tabular}
    \caption{Task accuracies for the Bulgarian language.}
    \label{tab:Bulgarian}
\end{table*}

\begin{table*}[t]
    \centering
    \begin{tabular}{lccccc}
            \toprule
            Model & Average & EU21-ARC & EU21-HeSw & EU21-MMLU & EU21-TQA \\
            \midrule
            Meta-Llama-3.1-8B-Instruct & .566 & .570 & .576 & .572 & .546 \\
            Mistral-7B-Instruct-v0.3 & .556 & .579 & .561 & .506 & .577 \\
            Meta-Llama-3.1-8B & .556 & .569 & .585 & .552 & .516 \\
            Salamandra-7B-Instruct & .547 & .592 & .634 & .460 & .502 \\
            Aya-23-8B & .548 & .560 & .620 & .495 & .518 \\
            Mistral-7B-v0.3 & .536 & .568 & .553 & .519 & .506 \\
            Salamandra-7B & .525 & .595 & .632 & .409 & .464 \\
            Occiglot-7B-eu5-Instruct & .474 & .507 & .500 & .432 & .457 \\
            Occiglot-7B-eu5 & .458 & .486 & .489 & .426 & .430 \\
            Pharia-1-LLM-7B-ctr-aligned & .336 & .282 & .314 & .313 & .436 \\
            Bloomz-7B1 & .334 & .274 & .299 & .287 & .478 \\
            Pharia-1-LLM-7B-ctr & .332 & .275 & .316 & .297 & .440 \\
            Bloom-7B1 & .326 & .272 & .297 & .260 & .476 \\
            \midrule
            Teuken-7B-Instruct (Ours) & .576 & .589 & .656 & .456 & .603 \\
            Teuken-7B-Base (Ours) & .518 & .555 & .606 & .444 & .468 \\
            \bottomrule
    \end{tabular}
    \caption{Task accuracies for the Polish language.}
    \label{tab:Polish}
\end{table*}

\begin{table*}[t]
    \centering
    \begin{tabular}{lccccc}
            \toprule
            Model & Average & EU21-ARC & EU21-HeSw & EU21-MMLU & EU21-TQA \\
            \midrule
            Meta-Llama-3.1-8B-Instruct & .535 & .537 & .521 & .559 & .525 \\
            Salamandra-7B-Instruct & .530 & .581 & .613 & .462 & .465 \\
            Meta-Llama-3.1-8B & .523 & .522 & .531 & .542 & .496 \\
            Salamandra-7B & .507 & .586 & .611 & .411 & .421 \\
            Mistral-7B-Instruct-v0.3 & .476 & .468 & .468 & .475 & .491 \\
            Mistral-7B-v0.3 & .458 & .454 & .464 & .484 & .432 \\
            Aya-23-8B & .456 & .429 & .489 & .450 & .454 \\
            Occiglot-7B-eu5-Instruct & .422 & .418 & .438 & .412 & .420 \\
            Occiglot-7B-eu5 & .418 & .406 & .430 & .412 & .425 \\
            Bloomz-7B1 & .330 & .262 & .291 & .281 & .488 \\
            Pharia-1-LLM-7B-ctr-aligned & .326 & .264 & .302 & .313 & .426 \\
            Pharia-1-LLM-7B-ctr & .324 & .255 & .301 & .302 & .437 \\
            Bloom-7B1 & .316 & .263 & .294 & .255 & .453 \\
            \midrule
            Teuken-7B-Instruct (Ours) & .557 & .569 & .635 & .454 & .570 \\
            Teuken-7B-Base (Ours) & .505 & .537 & .591 & .451 & .442 \\
            \bottomrule
    \end{tabular}
    \caption{Task accuracies for the Slovak language.}
    \label{tab:Slovak}
\end{table*}

\begin{table*}[t]
    \centering
    \begin{tabular}{lccccc}
            \toprule
            Model & Average & EU21-ARC & EU21-HeSw & EU21-MMLU & EU21-TQA \\
            \midrule
            Meta-Llama-3.1-8B-Instruct & .513 & .510 & .489 & .539 & .513 \\
            Salamandra-7B-Instruct & .528 & .574 & .599 & .453 & .486 \\
            Mistral-7B-Instruct-v0.3 & .518 & .537 & .510 & .495 & .531 \\
            Salamandra-7B & .512 & .571 & .600 & .419 & .456 \\
            Meta-Llama-3.1-8B & .502 & .509 & .501 & .525 & .474 \\
            Mistral-7B-v0.3 & .503 & .529 & .506 & .501 & .476 \\
            Occiglot-7B-eu5-Instruct & .427 & .432 & .437 & .394 & .444 \\
            Occiglot-7B-eu5 & .420 & .418 & .434 & .393 & .437 \\
            Aya-23-8B & .391 & .334 & .379 & .395 & .457 \\
            Bloomz-7B1 & .319 & .242 & .292 & .275 & .467 \\
            Pharia-1-LLM-7B-ctr-aligned & .318 & .254 & .300 & .299 & .419 \\
            Pharia-1-LLM-7B-ctr & .317 & .259 & .300 & .295 & .413 \\
            Bloom-7B1 & .313 & .256 & .294 & .246 & .454 \\
            \midrule
            Teuken-7B-Instruct (Ours) & .551 & .557 & .615 & .452 & .580 \\
            Teuken-7B-Base (Ours) & .510 & .530 & .576 & .446 & .487 \\
            \bottomrule
    \end{tabular}
    \caption{Task accuracies for the Slovenian language.}
    \label{tab:Slovenian}
\end{table*}

\begin{table*}[t]
    \centering
    \begin{tabular}{lccccc}
            \toprule
            Model & Average & EU21-ARC & EU21-HeSw & EU21-MMLU & EU21-TQA \\
            \midrule
            Meta-Llama-3.1-8B-Instruct & .601 & .594 & .636 & .596 & .579 \\
            Mistral-7B-Instruct-v0.3 & .585 & .601 & .603 & .527 & .608 \\
            Meta-Llama-3.1-8B & .592 & .586 & .644 & .579 & .558 \\
            Salamandra-7B-Instruct & .566 & .620 & .664 & .459 & .519 \\
            Salamandra-7B & .543 & .608 & .660 & .428 & .477 \\
            Mistral-7B-v0.3 & .555 & .576 & .595 & .542 & .507 \\
            Occiglot-7B-eu5-Instruct & .510 & .512 & .548 & .452 & .526 \\
            Occiglot-7B-eu5 & .490 & .496 & .531 & .445 & .486 \\
            Aya-23-8B & .459 & .418 & .482 & .444 & .494 \\
            Pharia-1-LLM-7B-ctr-aligned & .378 & .324 & .352 & .360 & .476 \\
            Pharia-1-LLM-7B-ctr & .374 & .320 & .347 & .340 & .490 \\
            Bloomz-7B1 & .339 & .263 & .294 & .301 & .497 \\
            Bloom-7B1 & .332 & .275 & .298 & .257 & .498 \\
            \midrule
            Teuken-7B-Instruct (Ours) & .579 & .586 & .677 & .463 & .591 \\
            Teuken-7B-Base (Ours) & .527 & .552 & .627 & .458 & .473 \\
            \bottomrule
    \end{tabular}
    \caption{Task accuracies for the Swedish language.}
    \label{tab:Swedish}
\end{table*}

\subsubsection{Toxicity}

We evaluate our models compared to other instruction-tuned models on the PolygloToxicityPrompts benchmark \cite{jain2024polyglotoxicityprompts}.
We report results from $\textsc{PTP}_{\text{SMALL}}$. 
Our evaluation settings are the same as those detailed in Section 3.4 of \cite{jain2024polyglotoxicityprompts}, except the repetition parameter $K$ which is set to $K=1$ in our setup. 
We further report toxicity and profanity metrics provided by the Perspective API \cite{leesperspectiveapi}. 
For both attributes $i \in \{\text{Profanity}, \text{Toxicity}\}$ we report the \textsc{Empirical Probability} denoted $\text{EP}_i$ as well as the \textsc{Average} denoted $\text{A}_i$
as defined in Section~3.4 of \cite{jain2024polyglotoxicityprompts}. 
Tables \ref{tab:ptp_english} to \ref{tab:ptp_spanish} show the results by language as well as aggregated, for the languages German, English, French, Italian and Spanish.

Teuken's low toxicity scores can be attributed to the combination of our extensive filtering process and instruction tuning. First, we removed harmful content using heuristic rules and perplexity scores from KenLM. 
The resulting model still indicated room for improvement in toxicity. 
Therefore, for the last part of the training data, consisting of two trillion tokens, we applied an additional machine learning-based filter specifically trained to detect adult content. 
Combined with subsequent instruction tuning, this significantly reduced Teuken’s overall toxicity.

\begin{table*}[t]
    \centering
    \begin{tabular}{lcccc}
        \toprule
        Model & $\textsc{EP}_{\textit{Profanity}}$ & $\textsc{A}_{\textit{Profanity}}$ & $\textsc{EP}_{\textit{Toxicity}}$ & $\textsc{A}_{\textit{Toxicity}}$ \\
        \midrule
        Meta-Llama-3.1-8B-Instruct & .162 & .184 & .188 & .235 \\
        Mistral-7B-Instruct-v0.3 & .186 & .188 & .214 & .236 \\
        Aya-23-8B & .212 & .216 & .235 & .262 \\
        Bloomz-7B1 & .122 & \textbf{.132} & .149 & .173 \\
        Occiglot-7B-eu5-Instruct & .175 & .184 & .193 & .226 \\
        Salamandra-7B-Instruct & .205 & .207 & .221 & .251 \\
        \midrule
        Teuken-7B-Instruct (Ours) & \textbf{.069} & .136 & \textbf{.083} & \textbf{.165} \\
        \bottomrule
    \end{tabular}
    \caption{$\textsc{PTP}_{\text{small}}^{\text{English}}$ evaluated on instruction-tuned models.}
    \label{tab:ptp_english}
\end{table*}

\begin{table*}[t]
    \centering
    \begin{tabular}{lcccc}
        \toprule
        Model & $\textsc{EP}_{\textit{Profanity}}$ & $\textsc{A}_{\textit{Profanity}}$ & $\textsc{EP}_{\textit{Toxicity}}$ & $\textsc{A}_{\textit{Toxicity}}$ \\
        \midrule
        Meta-Llama-3.1-8B-Instruct & .035 & .081 & .029 & .097 \\
        Mistral-7B-Instruct-v0.3 & .022 & \textbf{.066} & .022 & .086 \\
        Aya-23-8B & .039 & .085 & .030 & .098 \\
        Bloomz-7B1 & .017 & .049 & .020 & \textbf{.066} \\
        Occiglot-7B-eu5-Instruct & .054 & .097 & .045 & .116 \\
        Salamandra-7B-Instruct & .045 & .091 & .037 & .102 \\
        \midrule
        Teuken-7B-Instruct (Ours) & \textbf{.010} & .077 & \textbf{.009} & .079 \\
        \bottomrule
    \end{tabular}   
    \caption{$\textsc{PTP}_{\text{small}}^{\text{German}}$ evaluated on instruction-tuned models.}
    \label{tab:ptp_german}
\end{table*}

\begin{table*}[t]
    \centering
    \begin{tabular}{lcccc}
        \toprule
        Model & $\textsc{EP}_{\textit{Profanity}}$ & $\textsc{A}_{\textit{Profanity}}$ & $\textsc{EP}_{\textit{Toxicity}}$ & $\textsc{A}_{\textit{Toxicity}}$ \\
        \midrule
        Meta-Llama-3.1-8B-Instruct & .264 & .258 & .250 & .268 \\
        Mistral-7B-Instruct-v0.3 & .156 & .171 & .138 & .179 \\
        Aya-23-8B & .227 & .232 & .221 & .245 \\
        Bloomz-7B1 & \textbf{.127} & \textbf{.129} & \textbf{.119} & \textbf{.136} \\
        Occiglot-7B-eu5-Instruct & .301 & .281 & .275 & .272 \\
        Salamandra-7B-Instruct & .314 & .294 & .286 & .288 \\
        \midrule
        Teuken-7B-Instruct (Ours) & .140 & .195 & .122 & .185 \\
        \bottomrule
    \end{tabular}
    \caption{$\textsc{PTP}_{\text{small}}^{\text{French}}$ evaluated on instruction-tuned models.}
    \label{tab:ptp_french}
\end{table*}

\begin{table*}[t]
    \centering
    \begin{tabular}{lcccc}
        \toprule
        Model & $\textsc{EP}_{\textit{Profanity}}$ & $\textsc{A}_{\textit{Profanity}}$ & $\textsc{EP}_{\textit{Toxicity}}$ & $\textsc{A}_{\textit{Toxicity}}$ \\
        \midrule
        Meta-Llama-3.1-8B-Instruct & .153 & .188 & .170 & .230 \\
        Mistral-7B-Instruct-v0.3 & .111 & .145 & .135 & .189 \\
        Aya-23-8B & .135 & .176 & .151 & .215 \\
        Bloomz-7B1 & .063 & \textbf{.090} & .077 & \textbf{.119} \\
        Occiglot-7B-eu5-Instruct & .184 & .208 & .192 & .241 \\
        Salamandra-7B-Instruct & .170 & .194 & .175 & .227 \\
        \midrule
        Teuken-7B-Instruct (Ours) & \textbf{.047} & .134 & \textbf{.051} & .143 \\
        \bottomrule
    \end{tabular}
    \caption{$\textsc{PTP}_{\text{small}}^{\text{Italian}}$ evaluated on instruction-tuned models.}
    \label{tab:ptp_italian}
\end{table*}

\begin{table*}[t]
    \centering
    \begin{tabular}{lcccc}
        \toprule
        Model & $\textsc{EP}_{\textit{Profanity}}$ & $\textsc{A}_{\textit{Profanity}}$ & $\textsc{EP}_{\textit{Toxicity}}$ & $\textsc{A}_{\textit{Toxicity}}$ \\
        \midrule
        Meta-Llama-3.1-8B-Instruct & .259 & .268 & .227 & .266 \\
        Mistral-7B-Instruct-v0.3 & .220 & .228 & .193 & .225 \\
        Aya-23-8B & .275 & .277 & .243 & .269 \\
        Bloomz-7B1 & .172 & \textbf{.177} & .163 & \textbf{.177} \\
        Occiglot-7B-eu5-Instruct & .293 & .287 & .262 & .277 \\
        Salamandra-7B-Instruct & .296 & .295 & .261 & .282 \\
        \midrule
        Teuken-7B-Instruct (Ours) & \textbf{.139} & .204 & \textbf{.113} & .192 \\
        \bottomrule
    \end{tabular}
    \caption{$\textsc{PTP}_{\text{small}}^{\text{Spanish}}$ evaluated on instruction-tuned models.}
    \label{tab:ptp_spanish}
\end{table*}

\subsection{Instruction Tuning}

In the following, we provide more insights regarding our instruction tuned model.
Particularly, we give an overview of the training datasets that we employed for instruction tuning the model (\Cref{appendix:instruction_tuning_datasets}), and present additional evaluation results (\Cref{appendix:instruction_tuning_evaluation}).

\subsubsection{Training details - Hyperparameters}
\label{app:it_hyperparams}
We display the utilized hyperparameters in \Cref{table:model_hyperparameters_it}.

\begin{table}[ht]
\centering
\begin{tabular}{lll}
\toprule
\textbf{Hyper-Parameter}         & \textbf{SFT}   & \textbf{DPO} \\ 
\midrule

Epochs                              & 2         & 1         \\
Weight decay                        & 0.1       & 0.0      \\ 
Batch size                          & 128       & 128      \\ 
Warmup steps                        & 4300      & 50       \\ 
Learning rate                       & 1e-5      & 3.25e-6  \\ 
Learning rate schedule              & cosine    & cosine   \\ 
Optimizer                           & AdamW     & AdamW    \\ 
Adam Beta1                          & 0.9       & 0.9      \\ 
Adam Beta2                          & 0.95      & 0.95     \\ 
Max. Sequence length                & 2048      & 1024     \\
Beta (DPO)                          & NA        & 0.1      \\
\bottomrule
\end{tabular}
\caption{Hyperparameter configuration during post-training}
\label{table:model_hyperparameters_it}
\end{table}

\subsubsection{Instruction Tuning Datasets}\label{appendix:instruction_tuning_datasets}

\begin{table*}[t]
    \centering
    \begin{threeparttable}
        \begin{tabular}{llll}
            \toprule
            Dataset & Sample Size & Min Distance & Languages \\
            \midrule
            IFeval-like \citep{if-eval-like}  & 5000 & 0.0 & EN \\
            GSM8K \citep{cobbe2021training}  & 8000 (EN) / 5000 (DE) & 0.0 & EN, DE \\
            Winogrande \citep{sakaguchi2021winogrande} & 4000 (per language) & 0.0 & EN, DE \\
            ARC \citep{clark2018think} & 180 (per language) & 0.0 & EN, DE \\
            HellaSwag \citep{zellers2019hellaswag} & 2000 (per language) & 0.0 & EN, DE \\
            Sigma & 30000 (per language) & 0.1 & EN, DE \\
            Sigma Evolved & 30000 (per language) & 0.2 & EN, DE \\
            Sigma Everyday conversations & 2260 & 0.0 & EN \\
            OpenMath-instruct 2 \citep{toshniwal2024openmath2} & 20000 (EN) / 5000 (DE) & 0.0 & EN, DE \\
            Teuken Guide & 20000 (EN) / 5000 (DE) & 0.0 & EN, DE \\
            Teuken Self Awareness SFT & 1130 & 0.0 & EN, DE \\
            \bottomrule
        \end{tabular}
    \end{threeparttable}
    \caption{SFT datasets including sample size, minimum distance, and used languages.}
    \label{tab:it_sft_data_composition_new}
\end{table*}

\begin{table*}[t]
    \centering
    \begin{tabular}{llll}
        \toprule
        Dataset & Sample Size & Languages \\
        \midrule
        
        SauerkrautLM-Fermented  & 3000  & DE    \\
        dpo-mix-7k  & 7000 (EN) / 2500 (per other language)  & EN,DE,FR,IT    \\
        truthy-dpo  & 1016 (EN) / 500 (per other language)  & EN,DE,FR,IT    \\
        Winogrande & 1500  & EN \\
        ARC & 180  & EN \\
        HellaSwag & 750 & EN \\
        
        \bottomrule
    \end{tabular}
    \caption{DPO datasets including sample size and languages.}
    \label{tab:it_dpo_data_composition_new}
\end{table*}

\Cref{tab:it_sft_data_composition_new} and \Cref{tab:it_dpo_data_composition_new} provide an overview of the employed instruction tuning datasets and DPO datasets, respectively. 

\subsubsection{Instruction Tuning Evaluation Results} \label{appendix:instruction_tuning_evaluation}

\label{apx:it_eval_results}
\begin{figure*}
     \centering
     \begin{subfigure}[b]{0.495\textwidth}
         \centering
         \includegraphics[clip, trim=1.2cm 0.5cm 0.1cm 0.6cm, width=\textwidth]{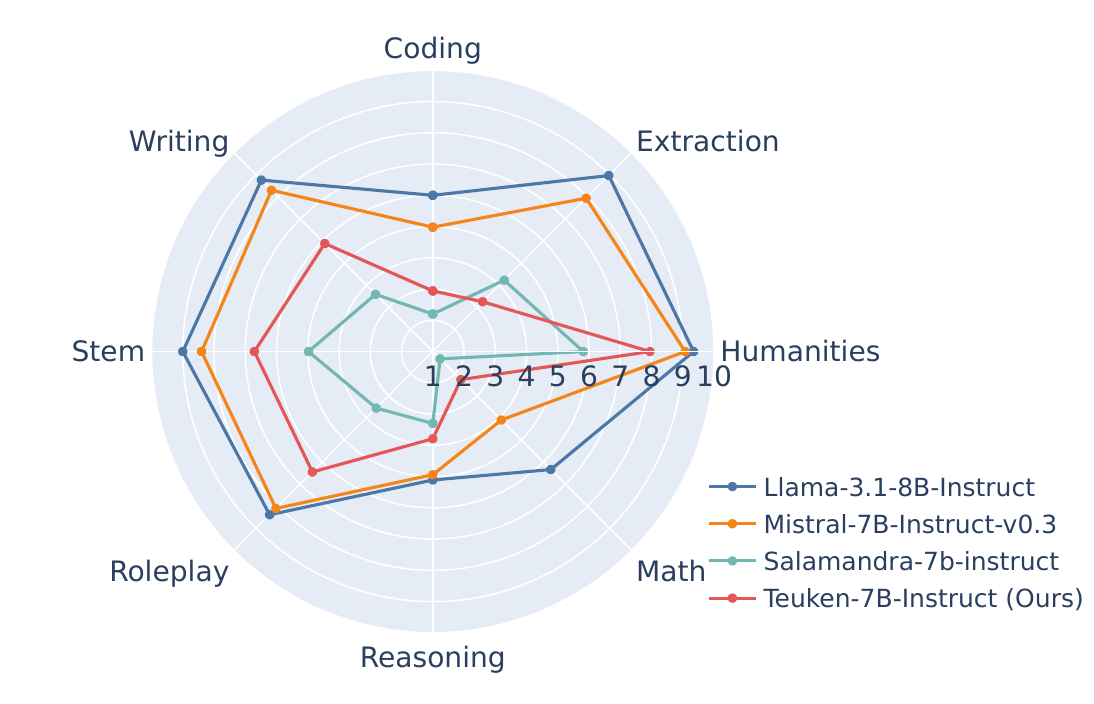}
        \caption{Cross-lingual avg. MT-Bench-X}
        \label{fig:radar-gpt-4-mt-bench-avg}
    \end{subfigure}    
    \hfill
     \begin{subfigure}[b]{0.495\textwidth}
         \centering
         \includegraphics[clip, trim=1.2cm 0.5cm 0.1cm 0.6cm, width=\textwidth]{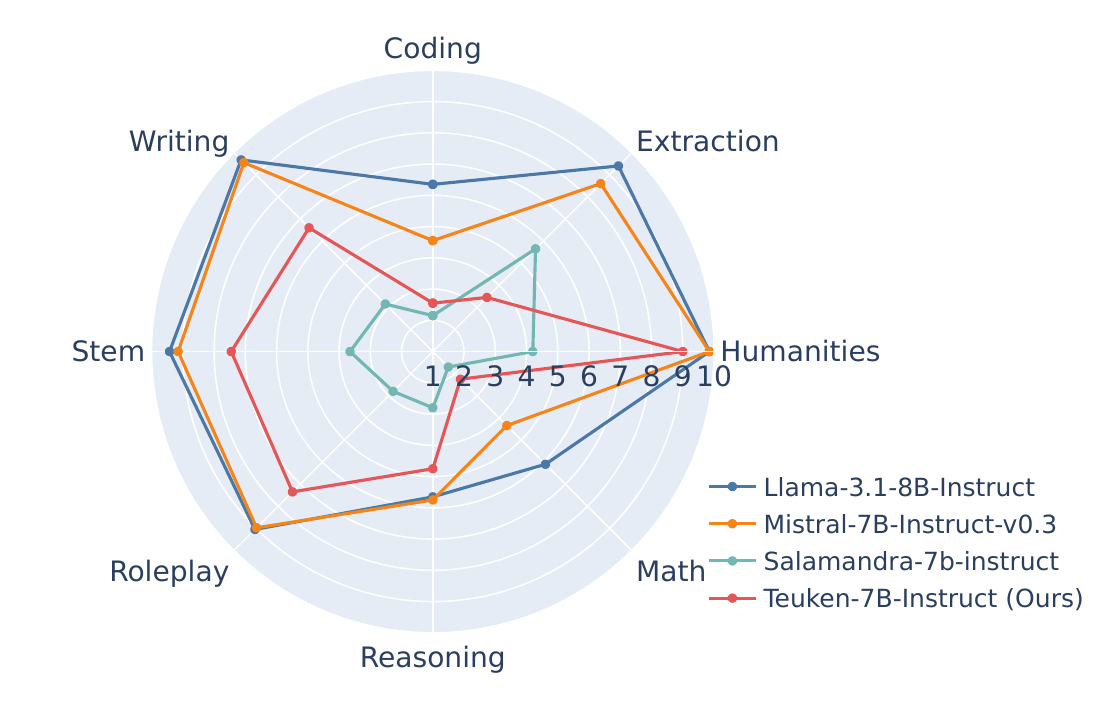}
        \caption{MT-Bench-EN}
        \label{fig:radar-gpt-4-mt-bench-EN}
    \end{subfigure}

     \par\vspace{0.5cm}
    
    \begin{subfigure}[b]{0.495\textwidth}
         \centering
         \includegraphics[clip, trim=1.2cm 0.5cm 0.1cm 0.6cm, width=\textwidth]{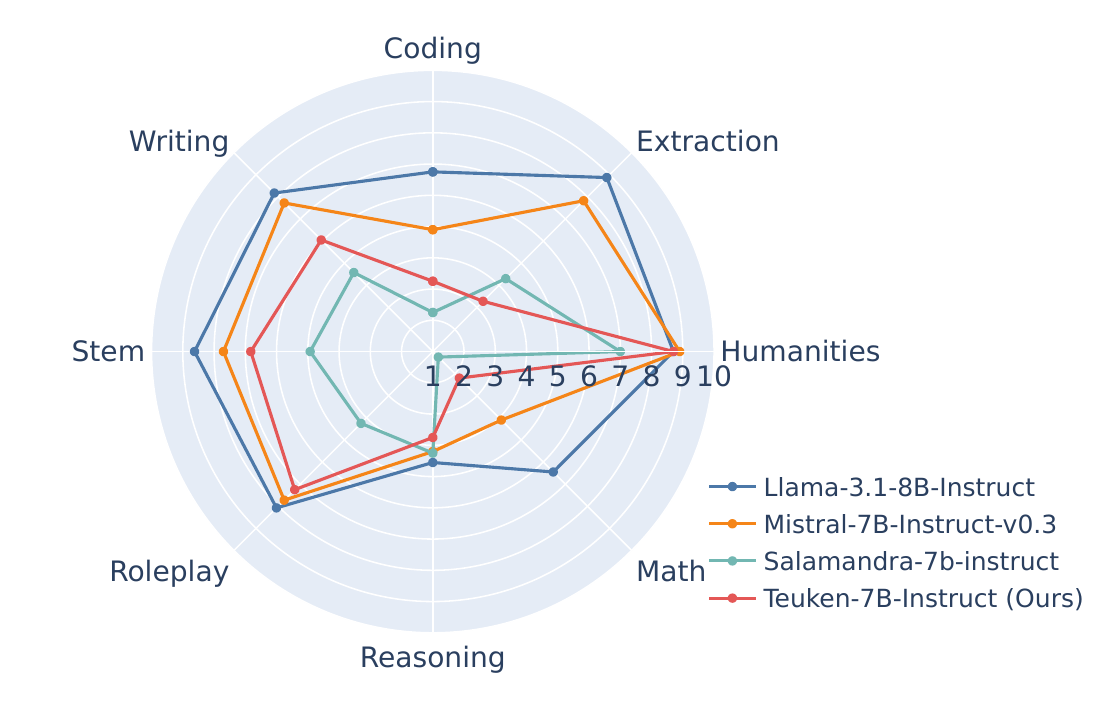}
        \caption{MT-Bench-DE}
        \label{fig:radar-gpt-4-mt-bench-DE}
    \end{subfigure}
    \hfill
    \begin{subfigure}[b]{0.495\textwidth}
         \centering
         \includegraphics[clip, trim=1.2cm 0.5cm 0.1cm 0.6cm, width=\textwidth]{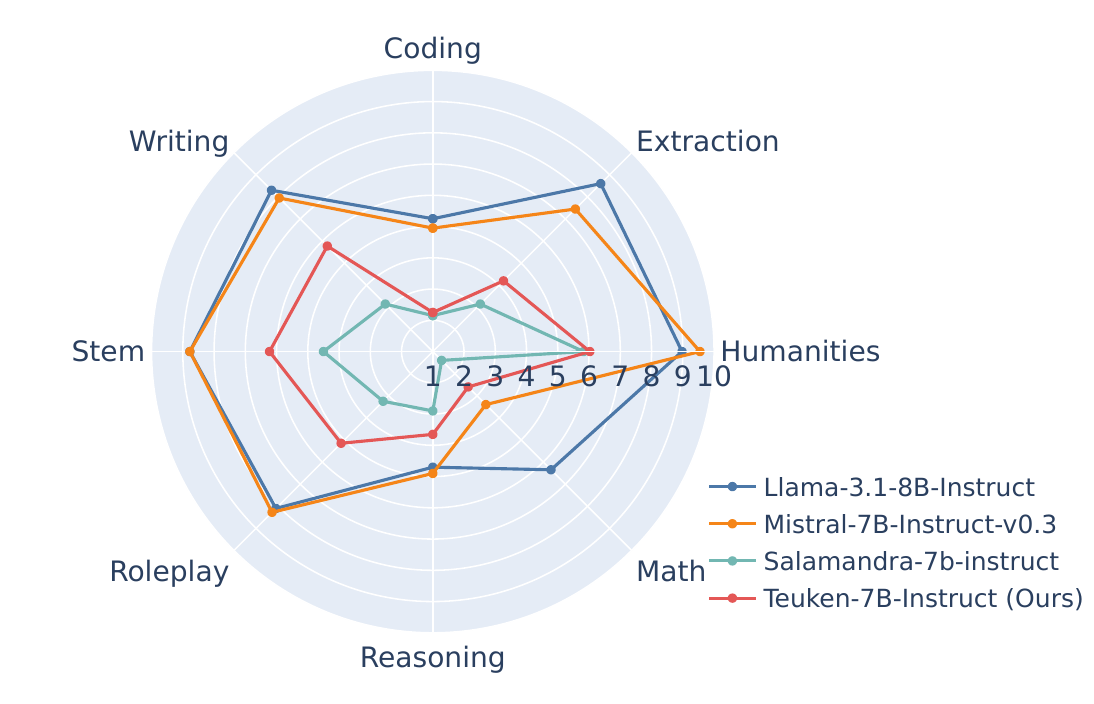}
        \caption{MT-Bench-FR}
        \label{fig:radar-gpt-4-mt-bench-FR}
    \end{subfigure}

     \par\vspace{0.5cm}
     
    \begin{subfigure}[b]{0.495\textwidth}
         \centering
         \includegraphics[clip, trim=1.2cm 0.5cm 0 0.1cm.6cm, width=\textwidth]{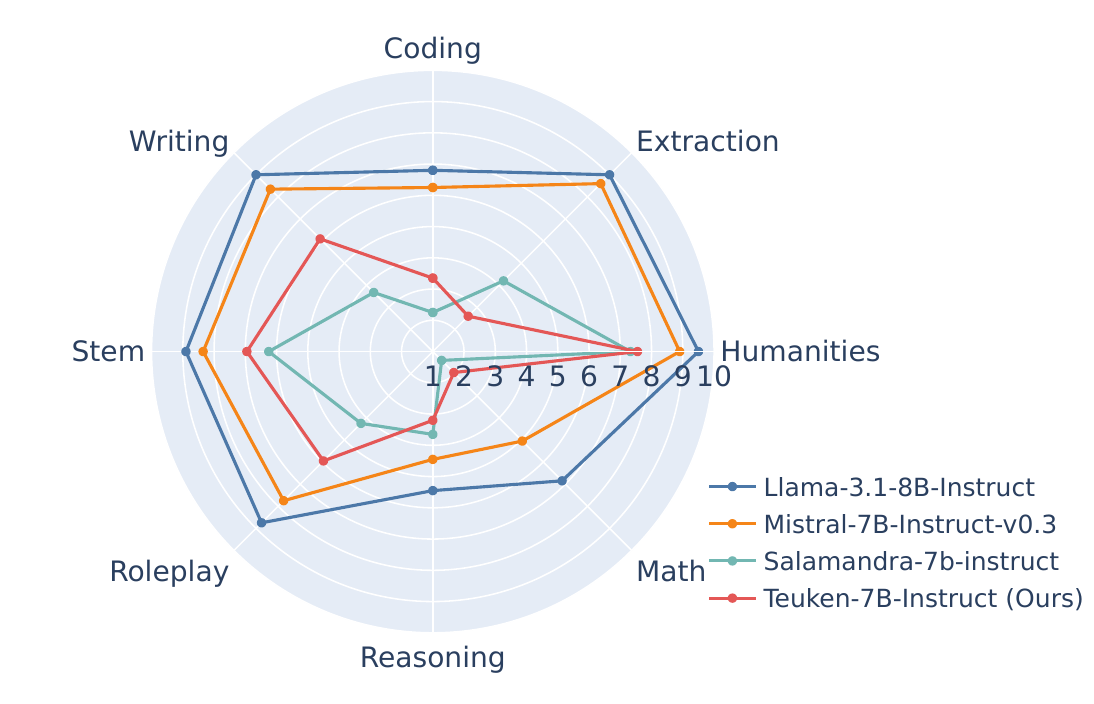}
        \caption{MT-Bench-IT}
        \label{fig:radar-gpt-4-mt-bench-IT}
    \end{subfigure}
    \hfill
    \begin{subfigure}[b]{0.495\textwidth}
         \centering
         \includegraphics[clip, trim=1.2cm 0.5cm 0.1cm 0.6cm, width=\textwidth]{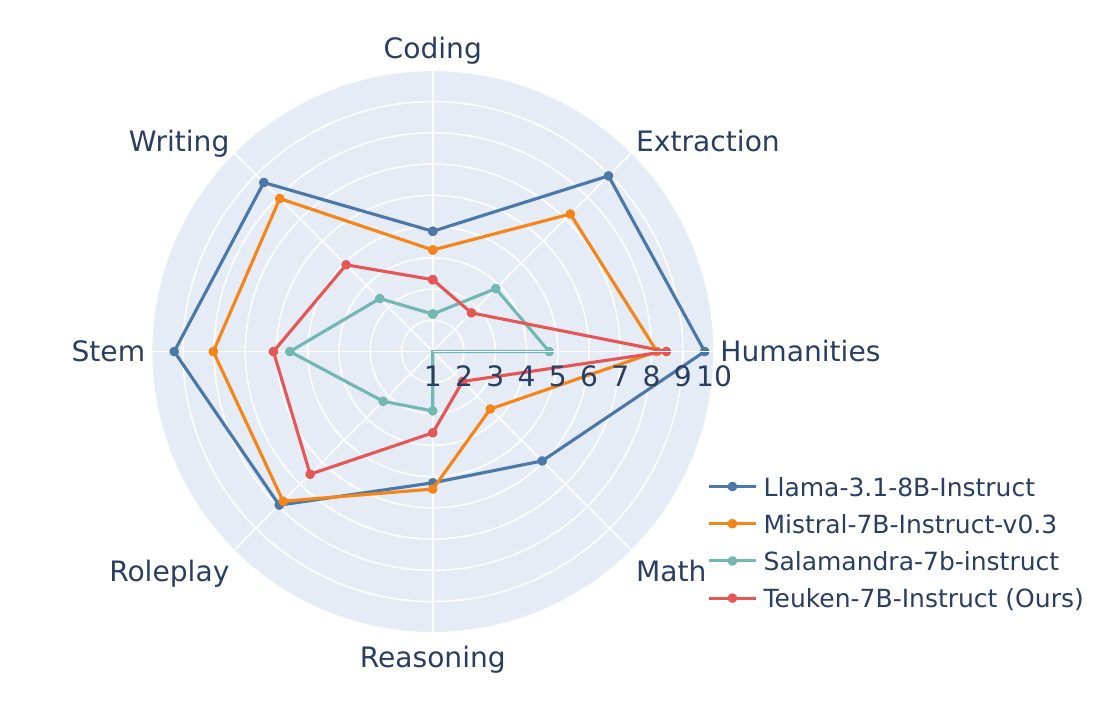}
        \caption{MT-Bench-ES}
        \label{fig:radar-gpt-4-mt-bench-ES}
    \end{subfigure}
     \par\vspace{0.5cm}
    \caption{In-depth MT-Bench-X quality assessment by GPT-4.}
    \label{fig:radar-gpt-4-mt-bench-X}
\end{figure*}

In \Cref{fig:radar-gpt-4-mt-bench-X}, we present additional evaluation results of MT-Bench-X~\cite{DBLP:journals/corr/abs-2402-13703}. 

\subsection{Software}\label{section:software}

We selected our training framework based on efficiency in terms of throughput (TFLOP/s) and maintenance.
Therefore, we decided to train our models based on a fork of Megatron-LM~\cite{korthikanti2023reducing} that supports various scalability features ensuring efficient training of transformer-based decoder-only models. Implementation contributions include a tensor-model-parallel SwiGLU layer~\cite{ramachandran2017searching,shazeer2020glu,raffel2020exploring,google2020t5v1.1} and fused RMSNorm~\cite{zhang2019root} integration.
In particular, we made use of 3D parallelism, i.e., data, tensor, and pipeline parallelism. 
Additionally, we used ZeRO~\cite{korthikanti2023reducing} to reduce memory requirements further by sharding the optimizer state.

\subsection{Training Infrastructure}\label{section:training_infrastructure}

We trained our models for 812.321 GPU hours on a supercomputer, which comprises 936 compute nodes, each containing 4$\times$ NVIDIA A100 (40 GB) GPUs connected via NVLink3 intra-node and through Mellanox HDR200 InfiniBand (IB) inter-node.

We utilized a maximum of 1024 GPUs and a minimum of 32 GPUs for our training runs, achieving perfect to near linear scaling behavior. Ablations were conducted to optimize the training runs for the system and network configuration in terms of parallelization, memory, and GPU utilization. The model employed tensor and pipeline parallelism of 2, along with sequence parallelism and optimizer sharding. State-of-the-art methods, such as flash attention and mixed-precision training, were also enabled, leading to maximum utilization of the hardware.

When scaling beyond 64 GPUs, the training became more susceptible to hardware failures. We encountered issues with high-bandwidth NCCL communications triggering hardware failures initiated by the IB port restarting. To enhance the robustness and fault tolerance of the training runs against node and port failures, we used environment variables for extending timeouts such as \texttt{NCCL\_IB\_TIMEOUT} and \texttt{UCX\_RC\_TIMEOUT}, along with \texttt{NCCL\_ASYNC\_ERROR\_HANDLING=1}. Furthermore, model checkpoints were saved at regular intervals and whenever errors or exit signals of the scheduler (Slurm) were detected.

The training runs were actively monitored using TensorBoard and custom cluster monitoring tool aiding to better visualization and debugging opportunities. We also implemented an automatic job submission strategy to combat hardware failures and queue times.

\clearpage
\onecolumn

\begin{table*}\centering\begin{tabular}{lrrrrr}\toprule{LAN} & {Curated (M)} & {Web (M)} & {Curated in \%} & {Total (M)} & {\%} \\\midrule\textbf{bg} & 21,241 & 23,827 & 47.13 & 60,803 & 1.13 \\\textbf{cs} & 6,231 & 46,950 & 11.72 & 73,724 & 1.33 \\\textbf{da} & 6,223 & 18,269 & 25.41 & 33,149 & 0.61 \\\textbf{de} & 36,865 & 312,216 & 10.56 & 470,396 & 8.73 \\\textbf{el} & 21,086 & 40,521 & 34.23 & 83,136 & 1.54 \\\textbf{en} & 212,905 & 1,453,724 & 12.77 & 2,787,190 & 41.67 \\\textbf{es} & 23,797 & 296,048 & 7.44 & 443,002 & 8.00 \\\textbf{et} & 8,788 & 6,258 & 58.41 & 19,356 & 0.38 \\\textbf{fi} & 3,236 & 36,020 & 8.24 & 54,749 & 0.98 \\\textbf{fr} & 30,617 & 333,625 & 8.41 & 487,932 & 9.11 \\\textbf{ga} & 474 & 75 & 86.26 & 699 & 0.01 \\\textbf{hr} & 15,302 & 1 & 99.99 & 18,894 & 0.38 \\\textbf{hu} & 3,378 & 37,574 & 8.25 & 56,178 & 1.02 \\\textbf{it} & 19,842 & 169,183 & 10.50 & 246,422 & 4.73 \\\textbf{lt} & 1,991 & 9,147 & 17.87 & 15,522 & 0.28 \\\textbf{lv} & 1,981 & 5,687 & 25.83 & 10,754 & 0.19 \\\textbf{mt} & 3,517 & 0.5 & 99.99 & 4,337 & 0.09 \\\textbf{nl} & 6,790 & 125,191 & 5.14 & 186,390 & 3.30 \\\textbf{pl} & 9,400 & 67,404 & 12.24 & 105,835 & 1.92 \\\textbf{pt} & 6,673 & 136,378 & 4.66 & 202,858 & 3.58 \\\textbf{ro} & 4,526 & 26,006 & 14.82 & 42,806 & 0.76 \\\textbf{sk} & 40,130 & 11,170 & 78.23 & 65,600 & 1.28 \\\textbf{sl} & 12,603 & 1,229 & 91.12 & 17,322 & 0.35 \\\textbf{sv} & 4,080 & 40,096 & 9.24 & 61,333 & 1.10 \\\textbf{code} & 301,726 & - & 100.00 & 534,170 & 7.54 \\\midrule\textbf{Total} & 803,401 & 3,196,599 & 100.00 & 6,082,559 & 100.00 \\\bottomrule\end{tabular}\caption{Comparison of token counts (in millions) between curated and web data across different languages. The total word count reflects the sum of both curated and web-sourced data after deduplication and filtering. Source code is treated separately from natural language data.}\label{tab:combined_words}\end{table*}

\end{document}